\documentclass{article}
\usepackage{arxiv}
\usepackage{graphicx}
\usepackage[utf8]{inputenc} 
\usepackage{booktabs}       
\usepackage{amsfonts}       
\usepackage{nicefrac}       
\usepackage{microtype}      
\usepackage{lipsum}   
\usepackage[utf8]{inputenc}
\usepackage{algorithm} 
\usepackage{algpseudocode} 
\usepackage{xcolor}
\usepackage{hyperref}
\usepackage{mwe}
\usepackage{amssymb}
\usepackage{amsmath}
\usepackage{subfig}
\usepackage{float}
\usepackage{hyperref}       
\usepackage{url}            
\usepackage{booktabs}       
\usepackage{amsfonts}       
\usepackage{nicefrac}       
\usepackage{microtype}      
\usepackage{wrapfig}
\usepackage{grffile}

\title{Persistent Neurons}

\author{Yimeng Min 
\thanks{Yimeng Min is an intern at Mila - Quebec AI Institute} \\
Computer Science Department\\
Cornell University\\
Ithaca, NY 14853\\
\texttt{ym499cornell.edu}\\
}
\date{}
\begin{document}

\maketitle
Neural networks (NN)-based learning algorithms are strongly affected by the choices of initialization and data distribution. 
Different optimization strategies have been proposed for improving the learning trajectory and finding  a better optima.
However, designing improved optimization strategies is a difficult task under the conventional landscape view.
Here, we propose persistent neurons, a trajectory-based strategy that optimizes the learning task using information from previous converged solutions.
More precisely, we utilize the end of trajectories and let the parameters explore new landscapes by penalizing the model from converging to the previous solutions under the same initialization.  Persistent neurons can be regarded as a stochastic gradient method with informed bias where individual updates are corrupted by deterministic error terms.
Specifically, we show that persistent neurons, under certain data distribution, is able to converge to more optimal solutions while initializations under popular framework find bad local minima.  
We further demonstrate that persistent neurons helps improve the model's performance under both good and poor initializations. 
We evaluate the full and partial persistent model and show it can be used to boost the performance on a range of NN structures, such as AlexNet and residual neural network (ResNet).








\section{Introduction}

Neural networks(NN)-based architectures become the dominant learning approaches for many tasks including image classification, speech recognition. 
These methods have been applied to many other domains like potential  molecules discovery \cite{hinton2012deep} \cite{krizhevsky2012imagenet} \cite{ma2015deep} \cite{lecun2015deep}.   For achieving state of art performance, deeper neural network architectures are used, starting LeNet-5 to AlexNet and residual neural networks(ResNet) \cite{krizhevsky2012imagenet} \cite{lecun1998gradient} \cite{he2016deep}.
Most NN-based algorithms rely on backward propagation\cite{rumelhart1985learning} to update the parameters in the network.   Back-propagating the gradient optimizes a training criterion with respect to a set of parameters $w$. Iterating the training process aims at finding the function $f({w})$ that minimizes some expected loss. Starting for initial parameter set $\boldsymbol{ w_0}$, the position of $w$ is updated every training iteration. The evolution history of $w$ corresponds to a trajectory in the parameter space.

Neural networks have highly non-convex loss surface and the number of local optima and saddle points can grow exponentially as the number of parameters increases \cite{auer1996exponentially}. At the same time, as the structure goes deeper, gradient vanishing/exploding exacerbates the learning.
These barriers hinder the trajectory converging to the optimal points for the expected loss.
A wide array of methods have been developed for improving the trainability of neural networks. Among all these methods, a good initialization is critical to achieve a desired functionality. Good weight initialization constitutes a favourable starting point in parameter space.  It also helps overcome the saturation problem and leads to more effective  training progress.  \cite{glorot2010understanding} \cite{erhan2010does}.
In most learning tasks, finding a global minimum for small network sizes can be NP-hard \cite{blum1989training}, and a proper initialization of the weights in a neural network is critical to the final convergence \cite{glorot2010understanding} \cite{he2015delving} \cite{mishkin2015all}.
\paragraph{Motivation}
{
In previous research \cite{glorot2010understanding} \cite{he2015delving}, the authors show good initialization leads to better performance. By monitoring the number of dead (or nearly dead) neurons, the authors also show that the model is less saturated with a better initialization. Instead of converging to local minima and stop updating, the parameters are more likely to reach the global optima or lower loss function values. However, recent studies show these initialization methods are not as solid as we think, and the definition of a good initialization is still ambiguous. Although the widely used initialization \cite{glorot2010understanding}\cite{he2015delving} methods show advantages on a range of tasks, there are also results suggest that the method proposed in \cite{he2015delving} strikingly fails on a shallow network\cite{holzmuller2020training}. A potential explanation is that people designed and compared different parameterizations by solely looking at the distribution of activations at the initialization stage instead of analyzing the trajectories of gradient descent (GD) \cite{he2015delving}\cite{mishkin2015all}. These initialization methods also do not consider different optimization strategies, but in practice, different gradient-based methods can lead to completely different results. This suggests that we should utilize more information from the trajectories.

\paragraph{Landscape Conjecture}
Minimizing the non-convex loss functions is central challenge for NN optimization. 
Non-convex error surfaces in high dimensional spaces generically suffer from a proliferation of saddle points \cite{dauphin2014identifying}. Many papers on optimization study the geometric of loss surface and reach a colloquial conjecture \cite{fyodorov2007replica}\cite{choromanska2015loss}\cite{ge2015escaping}\cite{ge2016matrix}\cite{kawaguchi2016deep}:
 \begin{center}
 \textbf{
No poor local minima — Local minima with high error are exponentially rare in high dimensions.
}
  \end{center}
For example, \cite{bray2007statistics} study the random Gaussian error functions on high dimensional continuous domains. They results suggest as the dimension becomes higher, it becomes exponentially unlikely to randomly pick all eigenvalues to be positive or negative, thus most critical points are saddle points. \cite{arora2018convergence}\cite{ge2015escaping}\cite{lee2016gradient} study the convergence on loss landscape. However, is the landscape conjecture applicable to the error landscapes of practical problems of interest? 
\paragraph{Why Trajectory-based? Limitations of the Landscape Conjecture}
It is hard to apply the landscape conjecture to deep networks. First, landscape conjecture proves convergence by disqualifying the poor local minimum and the non-strict saddle, but deep networks typically induce non-strict saddles\cite{kawaguchi2016deep} and the landscape perspective largely ignores algorithmic aspects that empirically are known to greatly affect convergence with deep networks — for example batch normalization\cite{ioffe2015batch}. 
Also, landscape conjecture tells us local minima with \textbf{high} error are exponentially rare. However, this does not mean all the minima have exact same accuracy. The solutions (minima) we find can still have small error gaps. The landscape conjecture fails to quantify the definition of \textbf{high} error. In real-world scenario, the minima are different and we are optimizing for a better minima (higher validation accuracy) in the loss surface. Moreover, recent study provides theoretical evdience that the error loss function presents \textbf{few} extremely wide flat minima (WFM) which coexist with narrower minima and critical points \cite{baldassi2020shaping}. This further strengthen the claim that the minima are different, in other words, not all the minima are the same--there exists good minima and poor minima. These results suggest that the landascape conjecture is too abstract to generlize towards practical problems of interest and we should use \textbf{trajectory-based} method for a better minima.


\paragraph{How Could Trajectory Help?}
Several papers have apply trajectory-based approach \cite{saad1995line}\cite{arora2018optimization}\cite{bartlett2018gradient}.
However, these above work use trajectory-based method in the context of linear model thus fail to generalize to real-world scenario. Designing trajectory-based method is especially difficult: different network structures can lead to different landscapes and converge to different solutions. Utilizing the right trajectory information is the key for generalization. 

In most cases, even with rigorous initialization strategies, parameters are still not placed at favourable locations. How to  keep optimizing and achieve the same advantages as better initialization by modelling the learning tasks using information from the trajectories? To be specific: instead of focusing on the start of the trajectories, by learning from other information of trajectories, can we improve the trainability and achieve advantages as follows?
\begin{itemize}
    \item If the network is born with poor initialization, can we design a method to optimize the trajectory to let the model has the same accuracy as a well-initialized  situation?
    \item Is the new trajectory capable of alleviating the gradient vanishing problem and not converging to sub-optimal minima or saddle points as a well-initialized situation?
\end{itemize}
Here, we propose Persistent Neurons. Contrary to initialization, we use the ends of the trajectories.

The statement \textit{using information from the ends of trajectories} seems counterintuitive since we are not able to know anything about the destination at the beginning.  As most learning tasks boil down about using deterministic descent method to reach particular minima, the results will keep the same if the initialization points remain unchanged. The model won't generate different solutions unless we use different initializations.
}

Persistent neurons is an approach for regularizing neural network using past optimization information and thus changes the gradient update during the training. As mentioned earlier, the model's parameters may converge to saddle points or local minima during gradient-based optimization. In persistent training, the weights start from the same initial point $\boldsymbol{w_{ini}}$ every time. We conjecture that the converged minima after the first training is not the global optimal point in the non-convex loss surface (Failure of \textit{no poor local minima} in landscape conjecture). To prevent the model converging to the same region, the loss term in the model includes additional penalties on the previous converged parameters recursively. The updated loss function is $f_n = f_{n-1} + g_{reg}(\boldsymbol{ w_{n-1}})$ where $f_{n-1}$ is the loss function of last persistent iteration and  $\boldsymbol{ w_{n-1}}$ is the converged parameters in the last training, respectively.
\paragraph{Contribution}
We summarize the contributions of this work as follows.
\begin{itemize}
    \item We propose a trajectory-based method, the method can be regarded as a gradient method with informed bias. To the best of our knowledge, this is the first attempt to add informed bias into gradient descent method.
    \item We show the persistent neurons can help model get rid of bad minima. We further introduce the full and partial persistent training and test on a range of structures. Our empirical results indicate that persistent neurons can improve generalization.
    \item We show that persistent neurons obtains better solution by investigating the hessian spectrum. We also analyse the neuron dynamics during training and found persistent neurons helps alleviate the neuron saturation problem.
\end{itemize}



\section{Persistent Training on Low Dimension}
\begin{figure}[H]
    \centering
    \begin{minipage}{0.5\textwidth}
        \centering
        \includegraphics[width=1.0\textwidth]{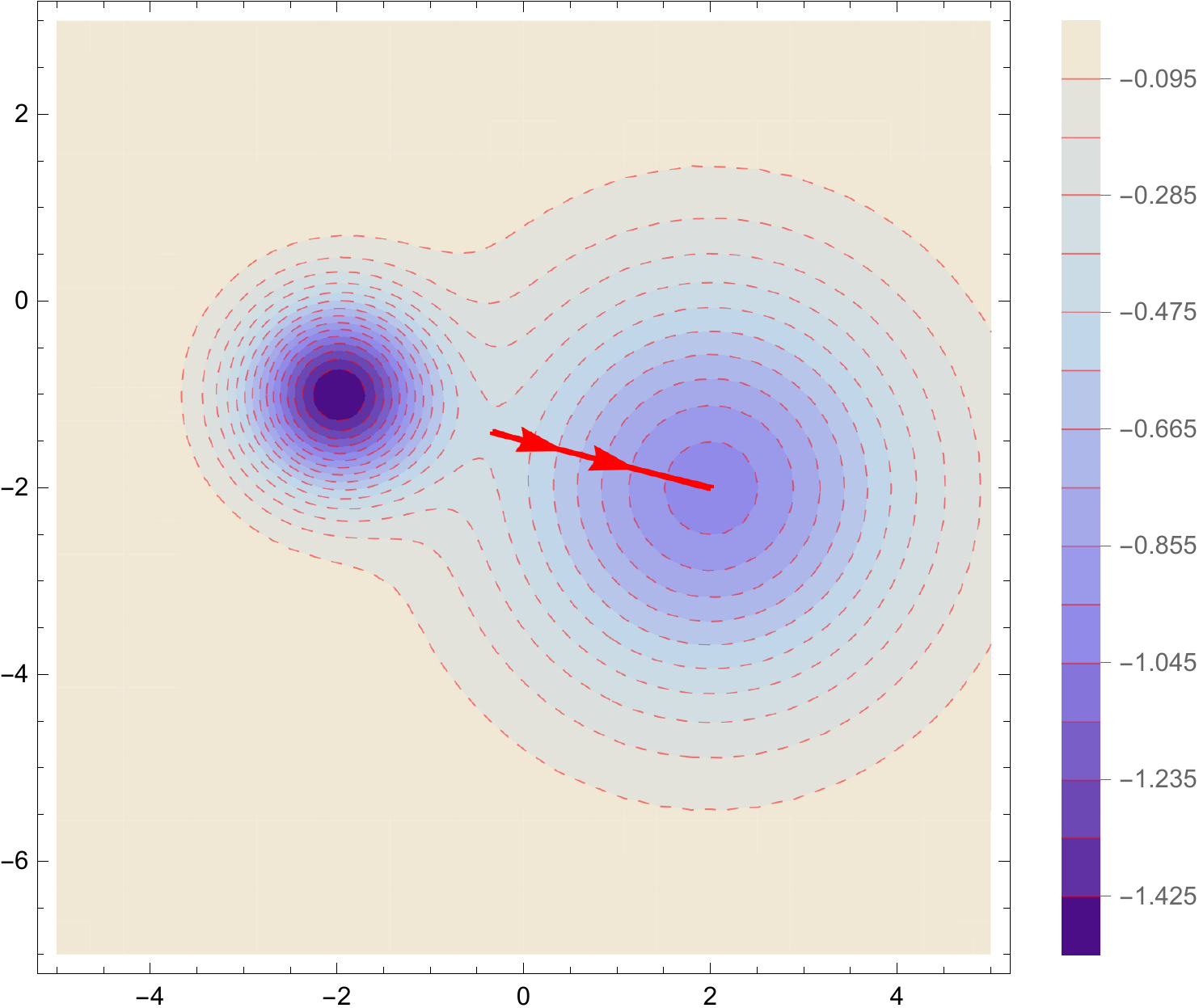} 
    \end{minipage}\hfill
    \begin{minipage}{0.5\textwidth}
        \centering
        \includegraphics[width=1.0\textwidth]{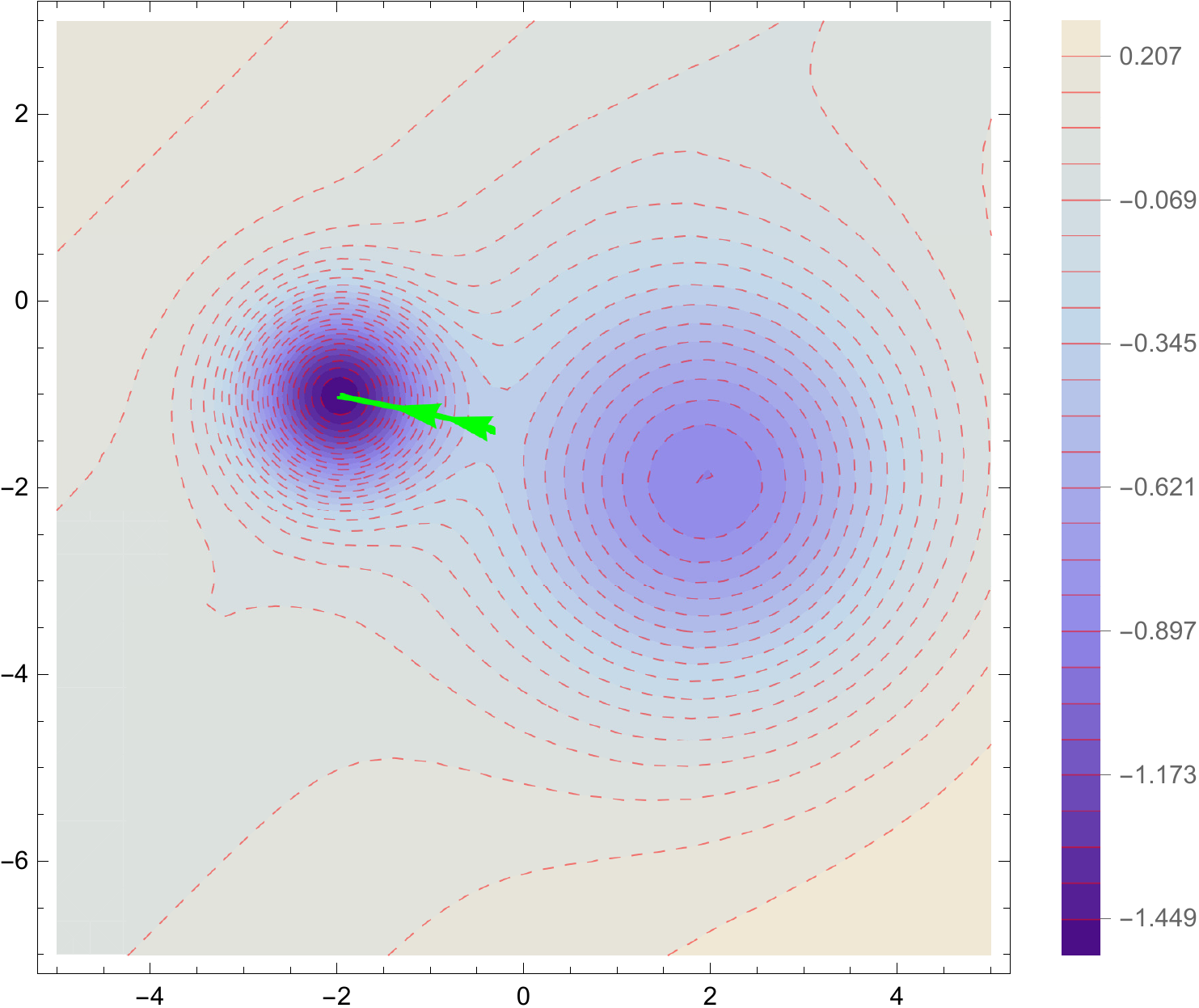} 
    \end{minipage}
    \caption{The optimization trajectory in the parameter space and the contour plot of the loss surface for plain model(left) persistent training(right). The learning rate is set as 0.001 and we run 50,000 steps GD.}
    \label{fig:2D_trac_0}
\end{figure}

Before empirically verifying the viability in deep neural networks, we start from a two-dimensional model for better visualizing. We use
gradient descent to update the parameters. We first define a loss function consists of two parameters: the function is written:
\begin{equation}
\begin{split}
    & f({w}) = f(x,y) = -\exp (-\frac{1}{5} ((x-2)^2+(y+2)^2))-\\
    & \frac{3}{2} \exp (-(x+2)^2+(y+1)^2)
\end{split}
\end{equation}{\label{eq:1}}
\begin{figure}[h]
\centering
\leavevmode
\centerline{\includegraphics[width=0.8\columnwidth]{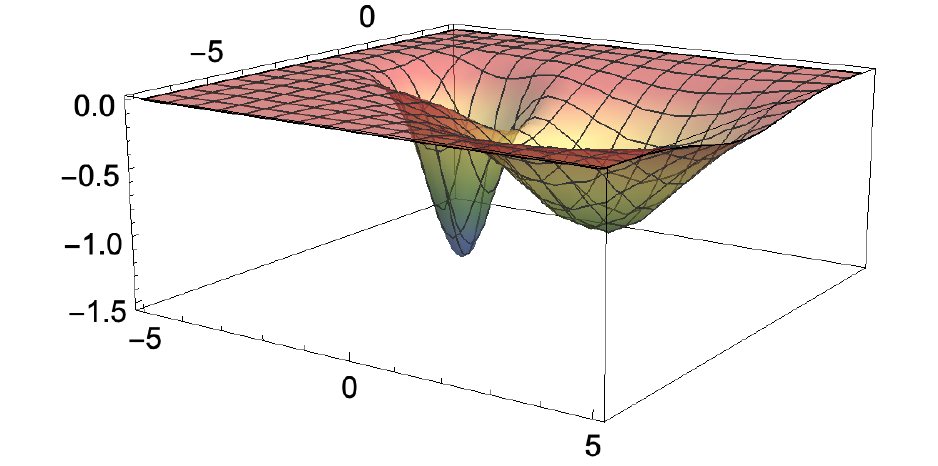}}\caption{Loss surface with two minima.}
\label{fig:3D_loss_surface}
\end{figure}
 $f$ has two minima: $\boldsymbol{ (x_a,y_a)} = (2,-2)$ and $\boldsymbol{ (x_b,y_b)} = (-2,-1)$ respectively.  The loss surface is shown in
 Figure~\ref{fig:3D_loss_surface}.
 There are two basins of attraction and the initialization of the weights is critical to its convergence. 
The initial parameter $\boldsymbol{w_{ini}}$ is set as (-0.335,-1.4), which locates in the middle between two basins of attraction. The gradient descent is: 
$$
x_{t+1} = x_t - \eta \nabla f({w})_x; \quad
y_{t+1} = y_t - \eta \nabla f({w})_y 
$$
where $t$ is the time step, $\eta$ is the learning rate and $\nabla f({ w})_{x,y}$ is the gradient along $x$ and $y$ directions. Figure~\ref{fig:2D_trac_0} shows the path and loss contour in the parameter space: after 50,000 steps gradient descent steps, $ w$ converges to the sub-optimal minima ($\boldsymbol{ x_a, y_a}$). 

We now apply persistent training to let $w$ discover another basin.  We add the previous converged parameters ${\boldsymbol{ w_0} = \boldsymbol{(x_a, y_a)}}$ into the iterative loss function. This eventually leads the new parameters getting rid of the attraction from the last converged basin. The updated loss function can be expressed as:
\begin{equation}
f_1 = f + g_{reg}(\boldsymbol{ w_0})
\end{equation}
where $g_{reg}(\boldsymbol{ w_0},{w}) = \lambda \times \frac{|\boldsymbol{w_0^T} { w}|}{||\boldsymbol{ w_0}||^2}$. $\lambda = 0.1 $ is hyperparameter that controls the penalty term on the previous converged weights. Smaller $\lambda$ requires more persistent training iterations.

Figure~\ref{fig:2D_trac_0} shows the loss surface and the optimizing trajectory of $f_1$. The weights descend from the same initial point $\boldsymbol{ w_{ini}}$ but converge to different basin(the global optima in this case). 
This iterative training method can be extended to more complex loss surfaces with more sub-optimal minima. The loss function during the $n_{th}$ iteration contains the regularization of all previous converged parameters $\boldsymbol{ w_{0,1,2,3...n-1}}$. By monitoring the validation performance, championship solution can thus be chosen. 

\begin{figure}[b]
        \centering
        \includegraphics[width=0.7\textwidth]{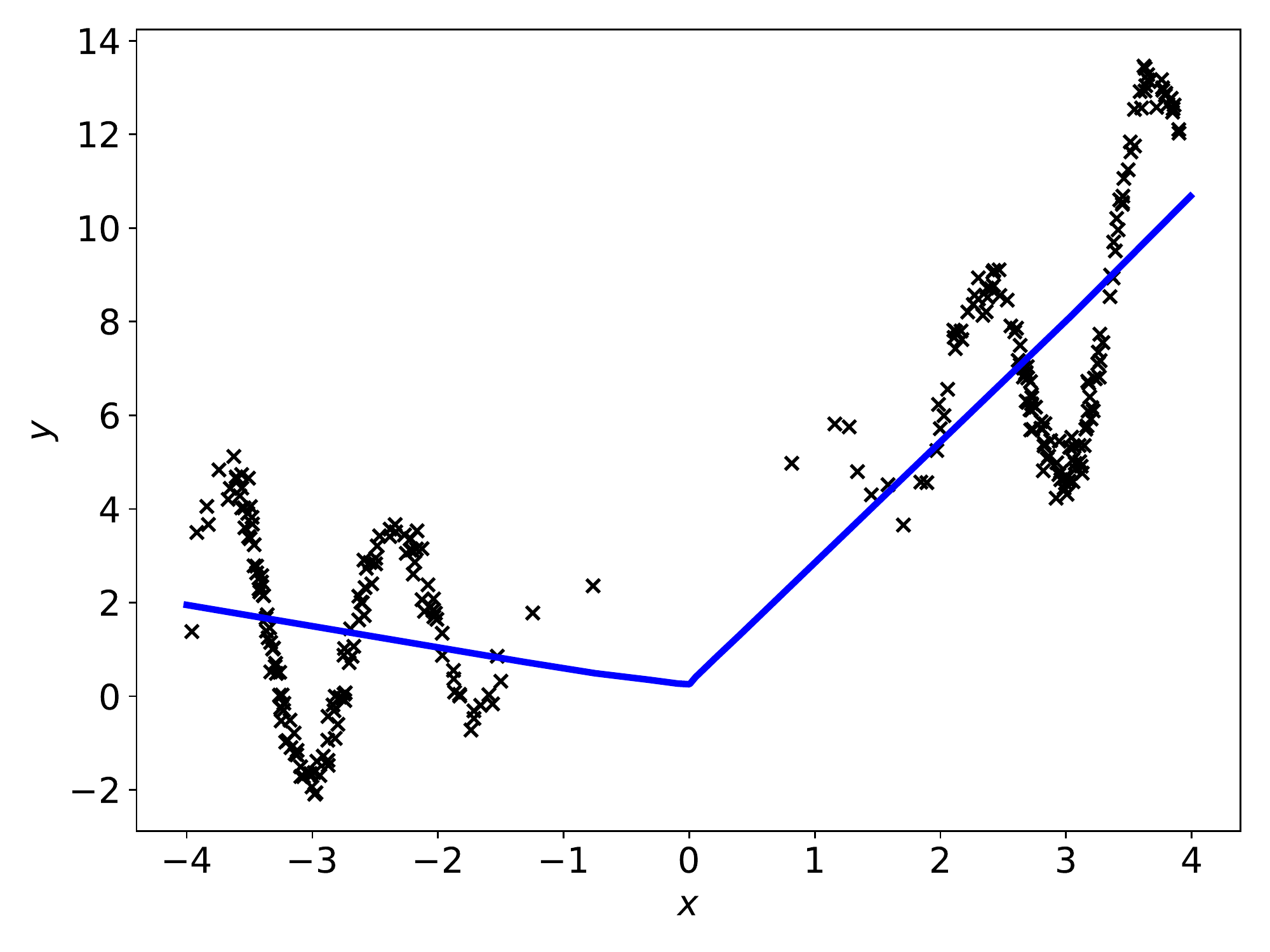} 
        \caption{NN predictor and the dataset, the predictor has a loss = 4.2 with a kink locates near zero.}
        \label{fig:NN_KINK}
\vspace{-6pt}
 \end{figure} 

\section{Persistent Training on Fully Connected Network}
In the last section, we showed that the parameters are able to get rid of the bad minima using persistent training. In the two-dimensional example, the initialization plays a deciding role in determining the final converged parameters. If we re-initialize the weights, for instance, assign the parameters to the left region in Figure~\ref{fig:2D_trac_0}, then GD is able to find the optima. In this section, we compare the re-initialization and persistent training. We show that the re-initialization under popular framework (He initialization for ReLU) fails in a simple shallow network and always produces similar results(degenerate functions in the parameter space) while persistent training is able to find different classes of function \cite{he2015delving}.

\begin{figure}[htb!]
    \centering
    \begin{minipage}{0.5\textwidth}
        \centering
        \includegraphics[width=0.9\textwidth]{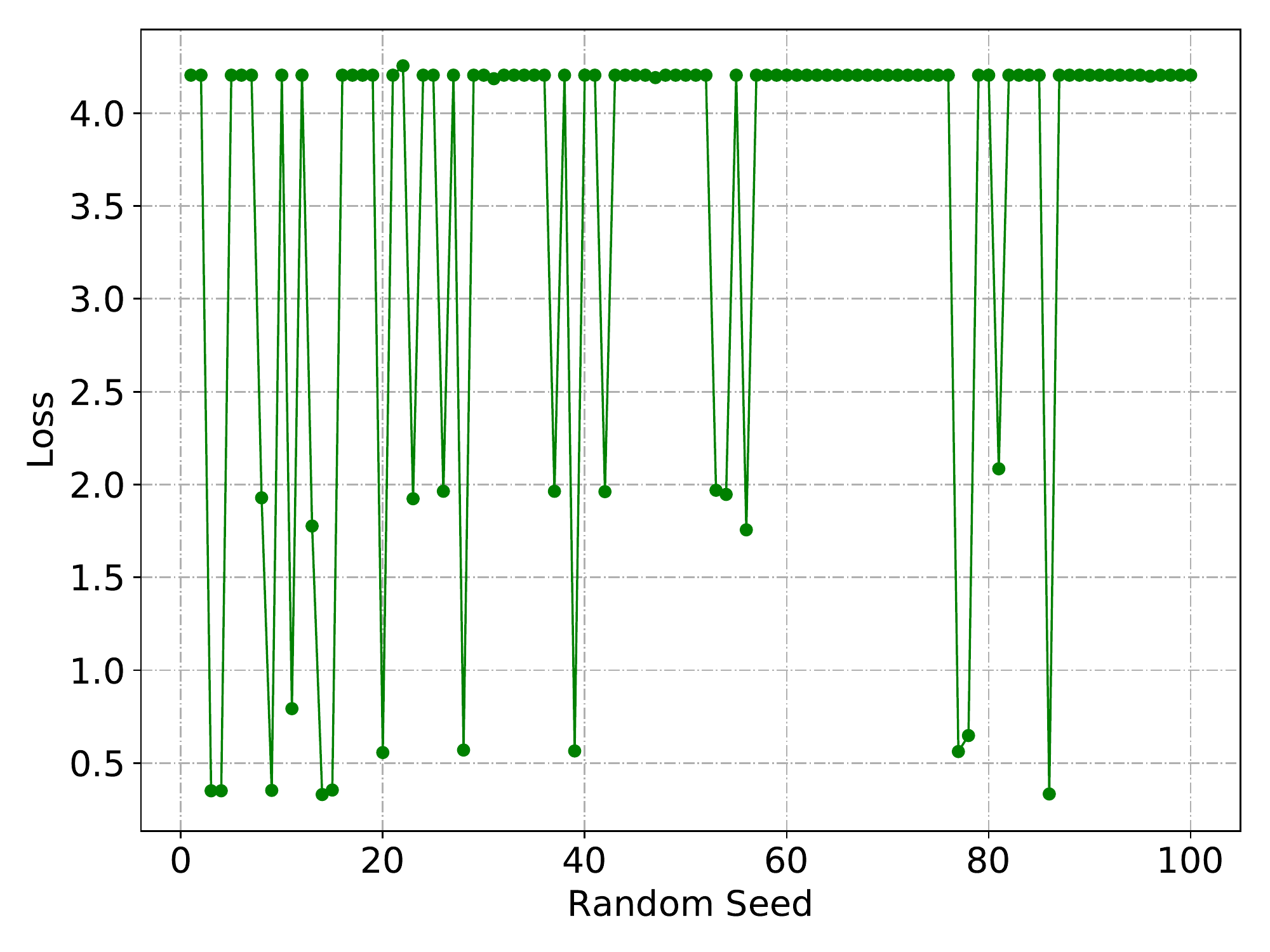} 
    \end{minipage}\hfill
    \begin{minipage}{0.5\textwidth}
        \centering
        \includegraphics[width=0.9\textwidth]{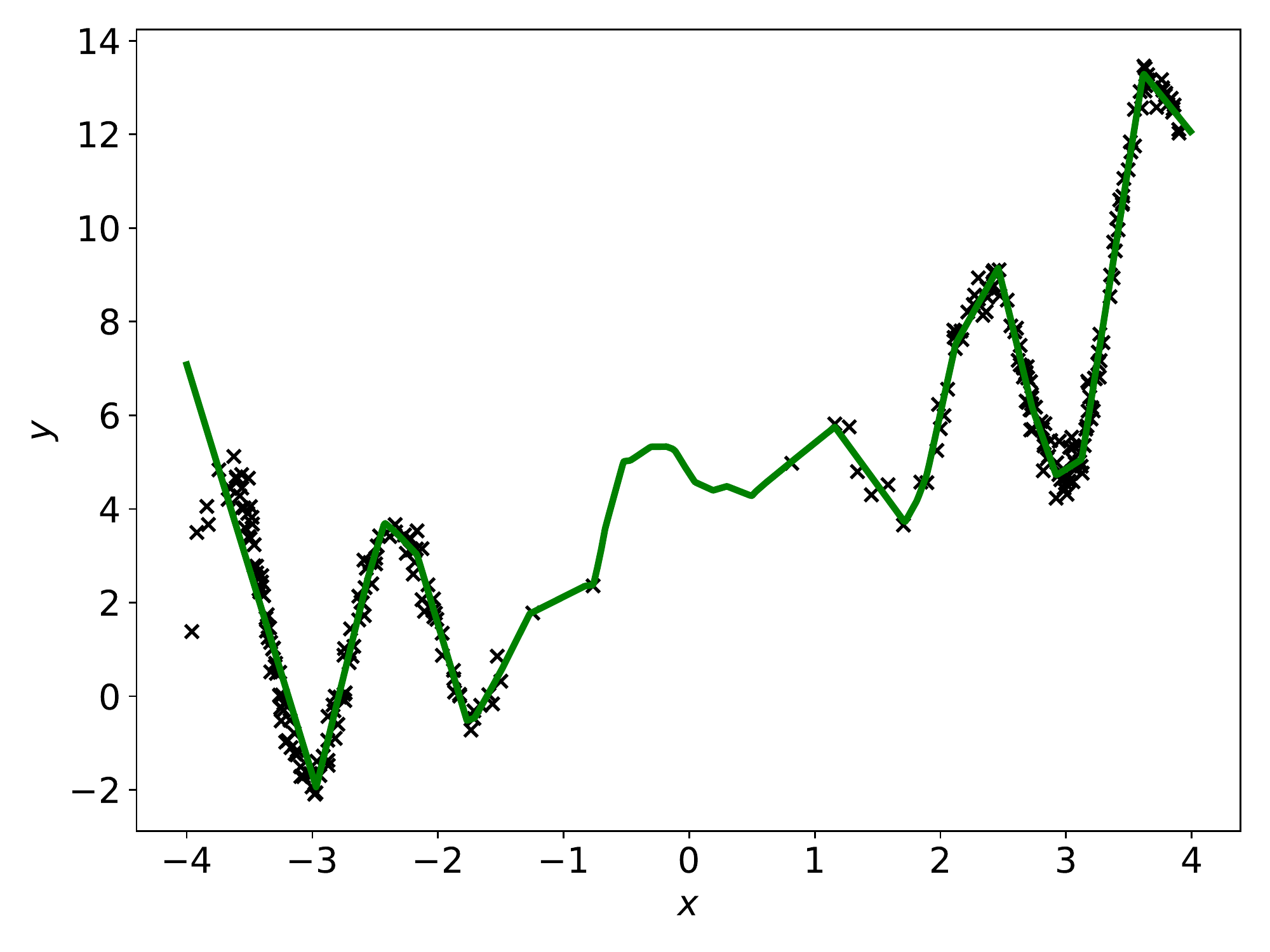} 
    \end{minipage}
    \caption{left:Loss history during 100  re-initializations using different random seeds, seed = 14 has the lowest loss; NN predictor's championship performance during re-initializations(corresponds to the lowest loss value model).}
   \label{fig:NN_predictor_0}
\end{figure}

\paragraph{Problem Setting}
Consider a three layer network with $m\in \mathbb{N}$  hidden units each layer.   $x,y \in \mathbb{R}$ and the data is shown in Figure~\ref{fig:NN_predictor_0}. We are interested in finding a function that fits the $y$  value. The three layer neural network defines a function($\mathbb{R} \rightarrow \mathbb{R}$):
\begin{equation}
f_{\theta}(x) = \boldsymbol{\theta}_3^T\sigma(\boldsymbol{\theta}_2 \sigma(\boldsymbol{\theta}_1 x +b_1) +b_2) +b_3
\label{eq:three_lay}
\end{equation}Where $\boldsymbol{\theta_3}\in \mathbb{R}^{m}$,$\boldsymbol{\theta_2}\in \mathbb{R}^{m\times m}$ and $\boldsymbol{\theta_1}\in \mathbb{R}^{m}$, $b_1,b_2,b_3$ are the bias terms and $\sigma$ is the ReLU activation function, the parameter at layer $i$ is the concatenation of $\theta_i$ and $b_i$. For the simplicity of notation, we will use $\theta_i$ to represents all parameters at layer $i$: $\theta_i$ $\leftarrow$ $\left[\theta_i \Vert b_i \right]$ in the following. We train parameters using GD with $momentum=0.9$  with mean square loss between the ground truth $y$ and the predicted value $f_\theta(x)$.   We first initialize the weight using the method proposed in \cite{he2015delving} and train the networks with 50000 steps. The result is shown in Figure~\ref{fig:NN_KINK}. We note that the model fails to fit the nonlinear trend and generate a kink(non-differential point).

\begin{algorithm}
	\caption{Training NN predictor with $N$ persistent iterations} 
	\begin{algorithmic}[1]
	\State \textbf{Input}: Initialization: $\boldsymbol{\Theta}_1^{ini}$,$\boldsymbol{\Theta}_2^{ini}$,$\boldsymbol{\Theta}_3^{ini}$, $\lambda = 0.01$, $m=32$,  persistent iterations: $N$
	\State \textbf{Output}: NN predictor  $f_{\theta}(x)$

	    	    \State Train the three layer model \hfill // {\color{blue} learning rate= 0.001, 50,000 iterations}
	    	    \State Save converged flattened parameters $\boldsymbol{\theta}_1$,$\boldsymbol{\theta}_2$,$\boldsymbol{\theta}_3$ as $\boldsymbol{ \Theta}_1^0$,$\boldsymbol{ \Theta}_2^0$,$\boldsymbol{ \Theta}_3^0$.  \hfill // {\color{blue} plain model training}
	    	  
	    	  \textit{\color{blue} persistent training starts here}
		\For{$iteration=1,2,\ldots N $} 

	    	        \State \textit{persistent-loss} =  $\sum_{i = 0}^{iteration - 1}$ ($\frac{ |{\boldsymbol{\Theta}_1^{i}}^T \boldsymbol{\theta}_1|}{|| \boldsymbol{\Theta}_1^i ||^2} $ +  $\frac{ |{\boldsymbol{\Theta}_2^{i}}^T \boldsymbol{\theta}_2|}{||\boldsymbol{\Theta}_2^i ||^2}$  +  $\frac{ |{\boldsymbol{\Theta}_3^{i}}^T \boldsymbol{\theta}_3|}{|| \boldsymbol{\Theta}_3^i ||^2} $ ) \hfill // \textit{\color{blue} persistent penalty} 
	    	        \State $\mathcal{L}_{persistent}$ = $\lambda$  $\times$ \textit{persistent-loss} + $\mathcal{L}$\hfill // {\color{blue} $\mathcal{L}$ is the vanilla loss}
	    	        \State Train  $f_\theta(x)$ w.r.t.  $\mathcal{L}_{persistent}$   \hfill // {\color{blue} learning rate= 0.001, 50,000 iterations}
	    	        \State Save converged flattened parameters as $\boldsymbol{\Theta}_1^{iteration}$,$\boldsymbol{\Theta}_2^{iteration}$,$\boldsymbol{\Theta}_3^{iteration}$. 
		\EndFor
	\State return $f_\theta(x)$ =  $f$($\boldsymbol{\Theta}_1^{N}$,$\boldsymbol{\Theta}_2^{N}$,$\boldsymbol{\Theta}_3^{N}$)($x$)
	\end{algorithmic} 
    \label{alg:1}
\end{algorithm}

For generating more reasonable predictors with different converged parameters, we first try re-initialization. During  100 times re-initializations, the frequency of observing the 'kink' in predicted functions is very high. Figure~\ref{fig:NN_predictor_0} shows the loss with $100$ times random initialization, we note that most random seeds generate trivial solutions (with loss $\sim$ 4.2), which suggests that the solutions remain affine linear on two different parts of the dataset. The model converges to bad local minima and performs linear regression in two regions even though the target $y$ is clearly very nonlinear.  Our observations suggest that the emergency of 'kink' is relatively robust when we randomly re-initialize the using the method proposed in \cite{he2015delving}. This problem has been reported by previous research\cite{holzmuller2020training}\cite{steinwart2019sober}. Figure~\ref{fig:NN_predictor_0} shows the championship predictor in 100 re-initializations, which corresponds to the lowest training loss of $0.33$.

\begin{figure}[H]
    \centering
    \begin{minipage}{0.5\textwidth}
        \centering
        \includegraphics[width=0.9\textwidth]{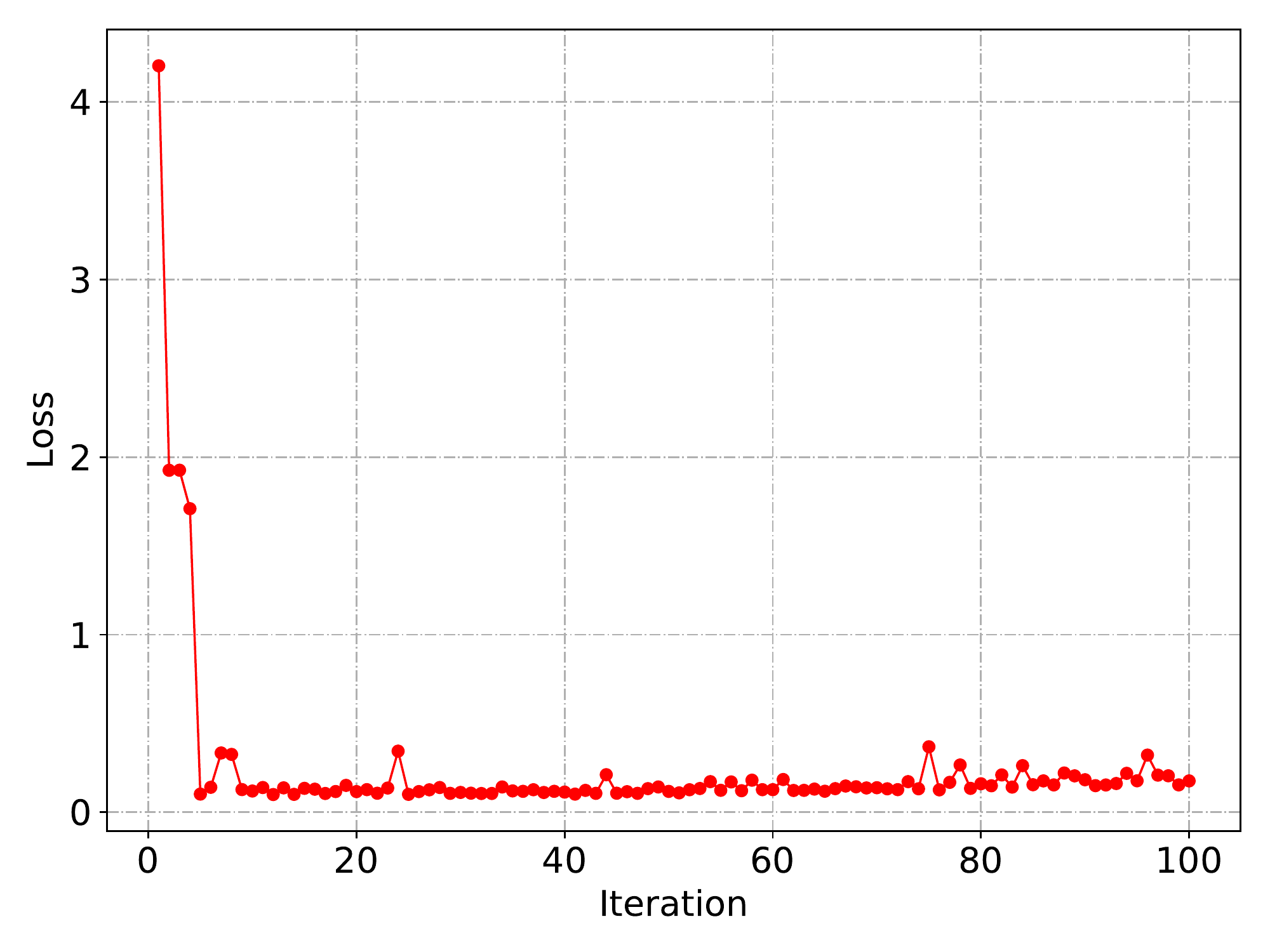} 
    \end{minipage}\hfill
    \begin{minipage}{0.5\textwidth}
        \centering
        \includegraphics[width=0.9\textwidth]{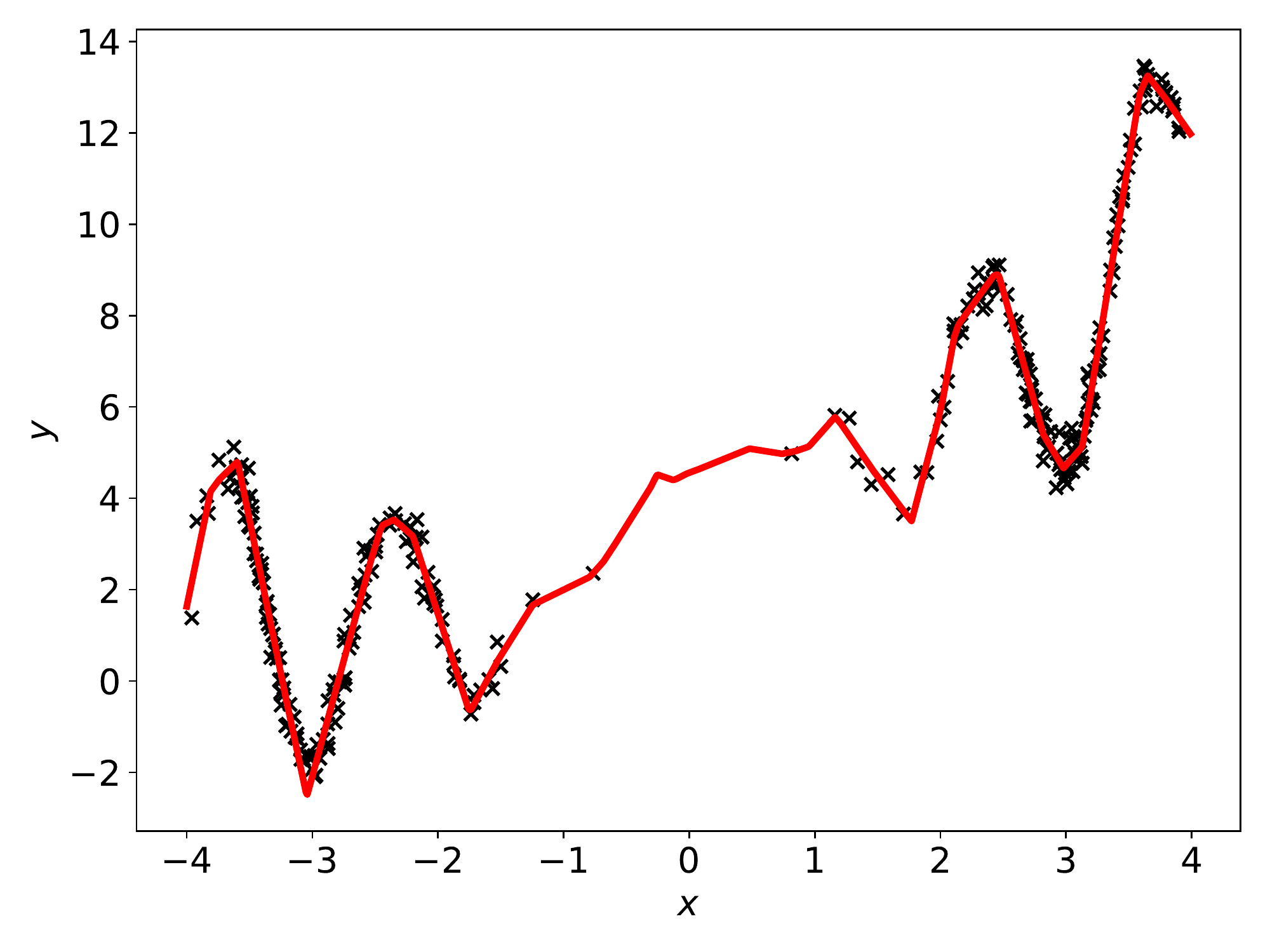} 
    \end{minipage}
    \caption{left:Loss history during 100 iterations  persistent training(1 plain model and 99 persistent training), $11_{th}$ persistent training has the lowest loss; right: NN predictor's championship performance during persistent training(corresponds to the lowest loss value model).}
   \label{fig:NN_predictor_loss_p}
\end{figure}

The failure of most random seeds implies that the initialization itself does not circumvent the parameters converging to sub-optimal solutions. Unlike the two dimensional cases in the previous section, re-initialization fails to improve the model's capability or to solve the 'kink' problem. These empirical results show that the commonly used initialization strategy does not help training under certain data distribution. 
Here, we use persistent training as an alternative strategy to solve this problem by utilizing information from previous converged trajectories. 
Algorithm~\ref{alg:1} shows the persistent training pseudocode for this fully connected network. We extract the previous converged parameters and add them as additional penalties to change the trajectories in the parameter space. Figure~\ref{fig:NN_predictor_loss_p} shows the corresponding loss using persistent training with $\lambda = 0.01$. 
We train the model for 100 iterations(1 plain model and 99 persistent iterations). 
Figure~\ref{fig:NN_predictor_loss_p} shows the championship predictor during persistent training, the championship model has loss value of $0.10$ while the 100 times random initializations' lowest loss is $0.33$. 
During persistent training, unlike re-initialization where the 'kink' arises repeatedly,  we observe that the non-differentiable point no longer exists after few training iterations($loss \ll 4.2$). 
The championship persistent predictor, though parameterized under unfavourable initialization, exhibits nonlinearity and  no longer performs linear regressions.  Figure~\ref{fig:NN_predictor_loss_p} shows that after several persistent training iterations,  the loss values maintain at a reasonable range instead of further increasing to $\sim 4.2$. This suggests that most persistent NN predictors no longer generate 'kink', and parameters in the shallow network avoid converging to bad minima. 
The results on shallow networks further strengthen the claim that persistent training is intrinsically different from re-initialization, which solely changes the start point of trajectory.



\section{Full and Partial Persistent Model}
In the previous section, we showed that persistent training helps improve the model's capability while re-initialization fails. 
This suggests that the method proposed in \cite{he2015delving}, though widely used, exhibits instability on certain data distribution. In the previous fully connected layer example, the initialization is a poor method. However, in many learning tasks, the method in  \cite{he2015delving} has proven to be a satisfying approach for initialization.  In these well-initialized models, can persistent training still improve the models' performance? In the following, we investigate persistent training with different models, including LeNet-5 to AlexNet and ResNet. 

\paragraph{Full Persistent Model}

The  loss for $n_{th}$ persistent training iteration is:
\begin{equation}
    \mathcal{L}_{n}({\theta}) = \mathcal{L}({\theta}) + \lambda   \sum_{k=0}^{n-1} \sum_{l=1}^m  \frac{{|\boldsymbol{\Theta}^l_{k}}^T{{\theta}^l}| }{ || \boldsymbol{\Theta}^l_k ||^2}
\end{equation}
Where $\mathcal{L}({\theta})$ is the network's loss function, $m$ is the number of layers in the network structure, ${ \theta =\left[\theta^1 \Vert \theta^2 \Vert...\Vert \theta^m \right] \in \mathbb{R}^{{\sum_l N_l}} }$ denotes the concatenation of parameters. $N_l$ is the number of parameters in layer $l$. $\theta_l$ is the flattened parameters at layer $l$. $\boldsymbol{ \Theta}^l_k$ corresponds to the flattened converged parameters at layer $l$ after $k_{th}$ persistent training, $\boldsymbol{\Theta}^l_0$ corresponds to plain model's converged parameters. $\lambda > 0$ is hyperparameter that controls the strength of persistent penalties.  
\begin{figure}[hbt!]
        \centering
        \includegraphics[width=0.9\textwidth]{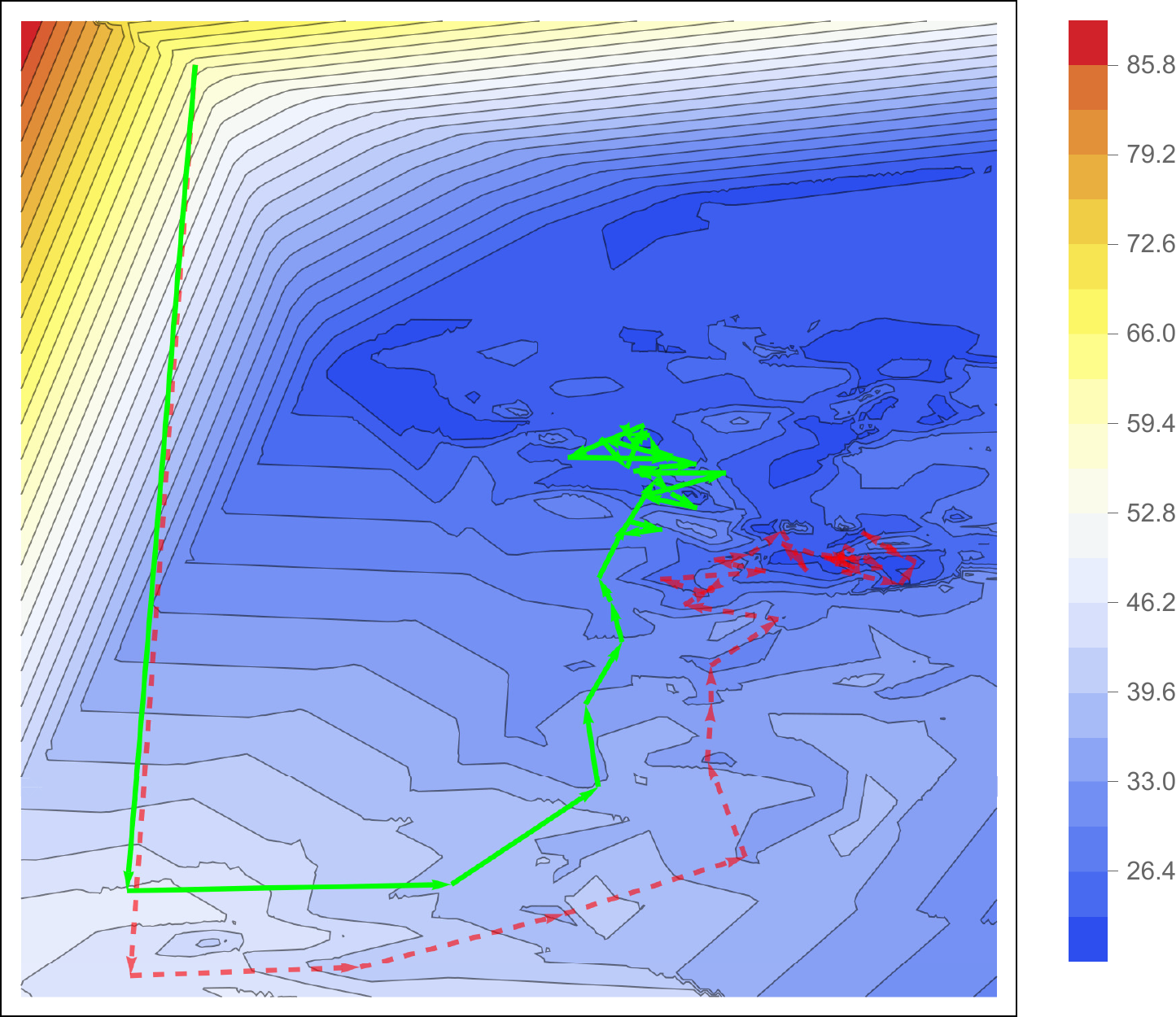} 
        \caption{Green: plain model training trajectory; Red: championship training trajectory.}
        \label{fig:LenetLosssurf}
\vspace{-6pt}   
 \end{figure}   
We sample the $\boldsymbol{\Theta}_{ini}$ from two different distributions: initialization 1,2 are sampled from normal distribution with zero mean and standard deviation 0.2 and 0.05, respectively, standing for a bad initialization and good one. We denote them as ${\boldsymbol{ \Theta_{ini_1}}}$ and ${\boldsymbol{ \Theta_{ini_2}}}$. 
We apply persistent training on LeNet-5 \cite{lecun1998gradient} with ReLU activation for classifying CIFAR-10 dataset. We use Adam optimizer \cite{kingma2014adam} with a batch size of 256.
The learning rate is 0.001 and the persistent penalty term $\lambda = 0.01$. We iteratively train the network for 20 times($1$ plain model and $19$ persistent models). There are 50000 images for training and 10000  images for testing in CIFAR-10, in our experiment, the original test data is randomly split into two datasets(5000 each) for validating and testing. 

Figure~\ref{fig:LenetLosssurf} shows the trajectories of plain model and persistent model. The two trajectories start from the same location while converge to different solutions as the persistent model includes the repulsion force from the previous converged basins. The full persistent model includes an orthogonal constraints on all  parameters.
We can also use the subset of the parameters as orthogonal constraints to prevent the model converge to same solutions, this partial persistent model also includes a repulsion force from the converged solution, which will be discussed in the next section.
\begin{figure}[t]
    \centering
    \begin{minipage}{0.5\textwidth}
        \centering
        \includegraphics[width=0.9\textwidth]{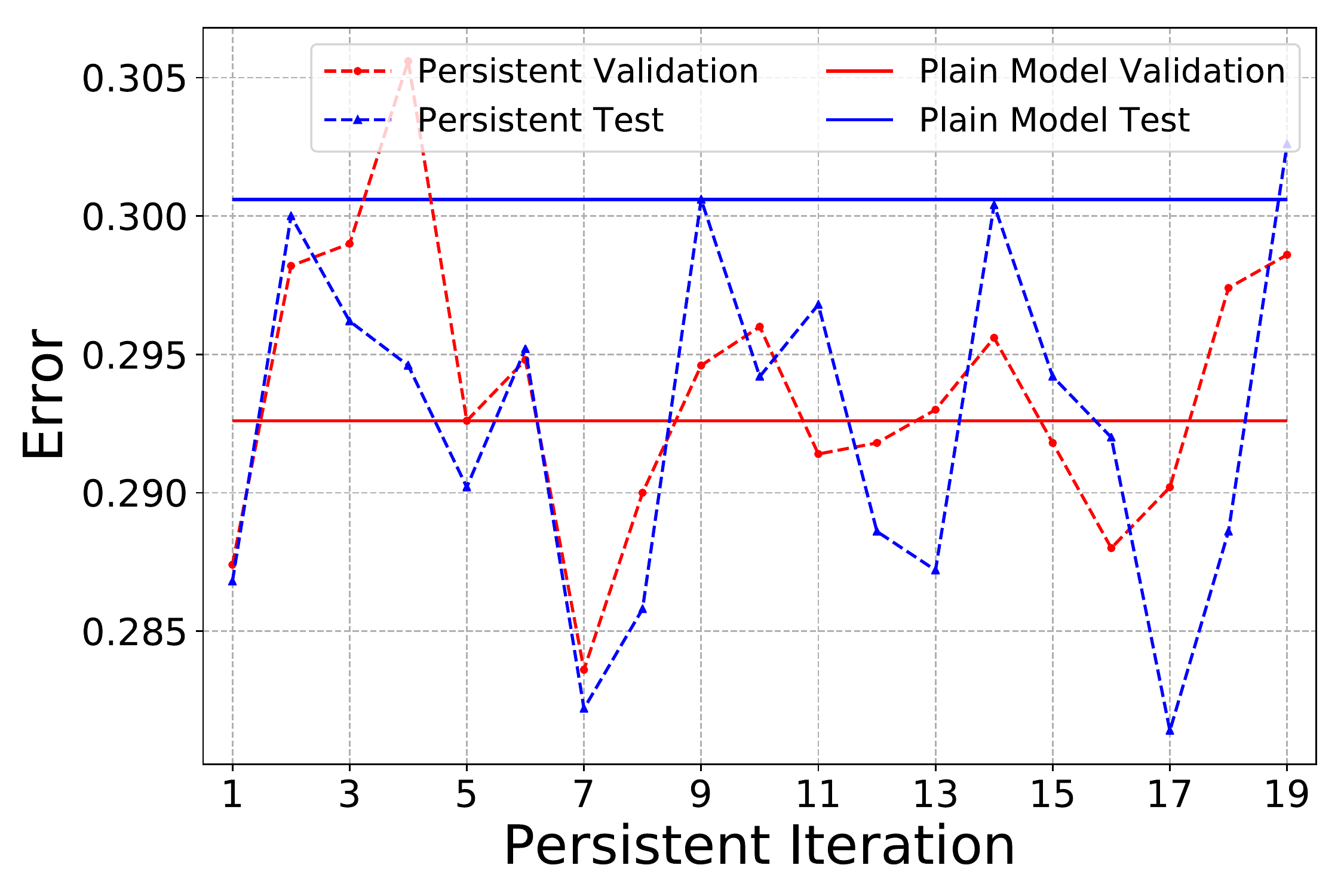} 
    \end{minipage}\hfill
    \begin{minipage}{0.5\textwidth}
        \centering
        \includegraphics[width=0.9\textwidth]{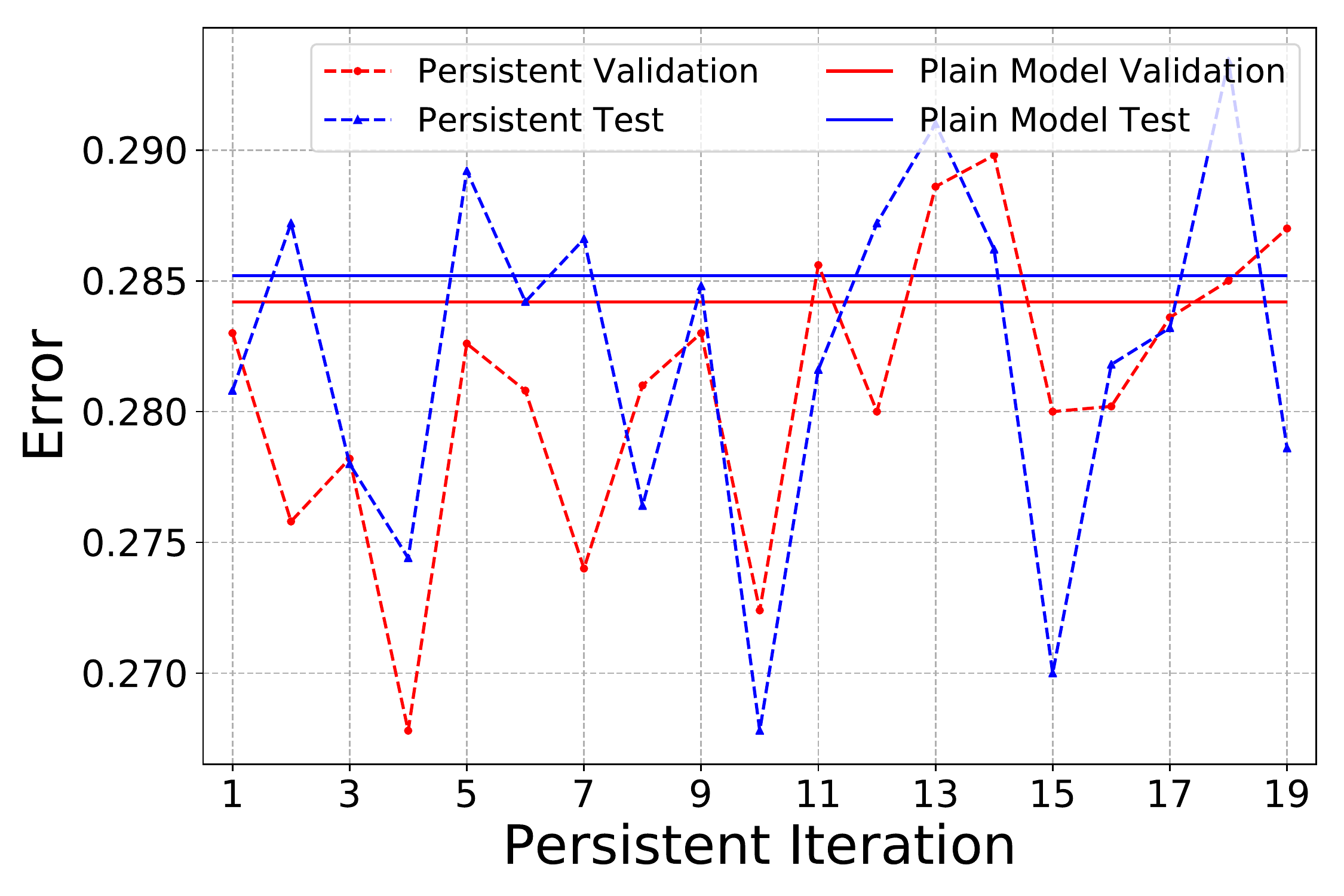} 
    \end{minipage}
    \caption{left: Validation and test error of  persistent iterations on CIFAR-10 (Initialization 1); right: Validation and test error of  persistent iterations on CIFAR-10 (Initialization 2).}
   \label{fig:lenet1}
\end{figure}
\begin{figure}[b]
    \centering
    \begin{minipage}{0.5\textwidth}
        \centering
        \includegraphics[width=0.9\textwidth]{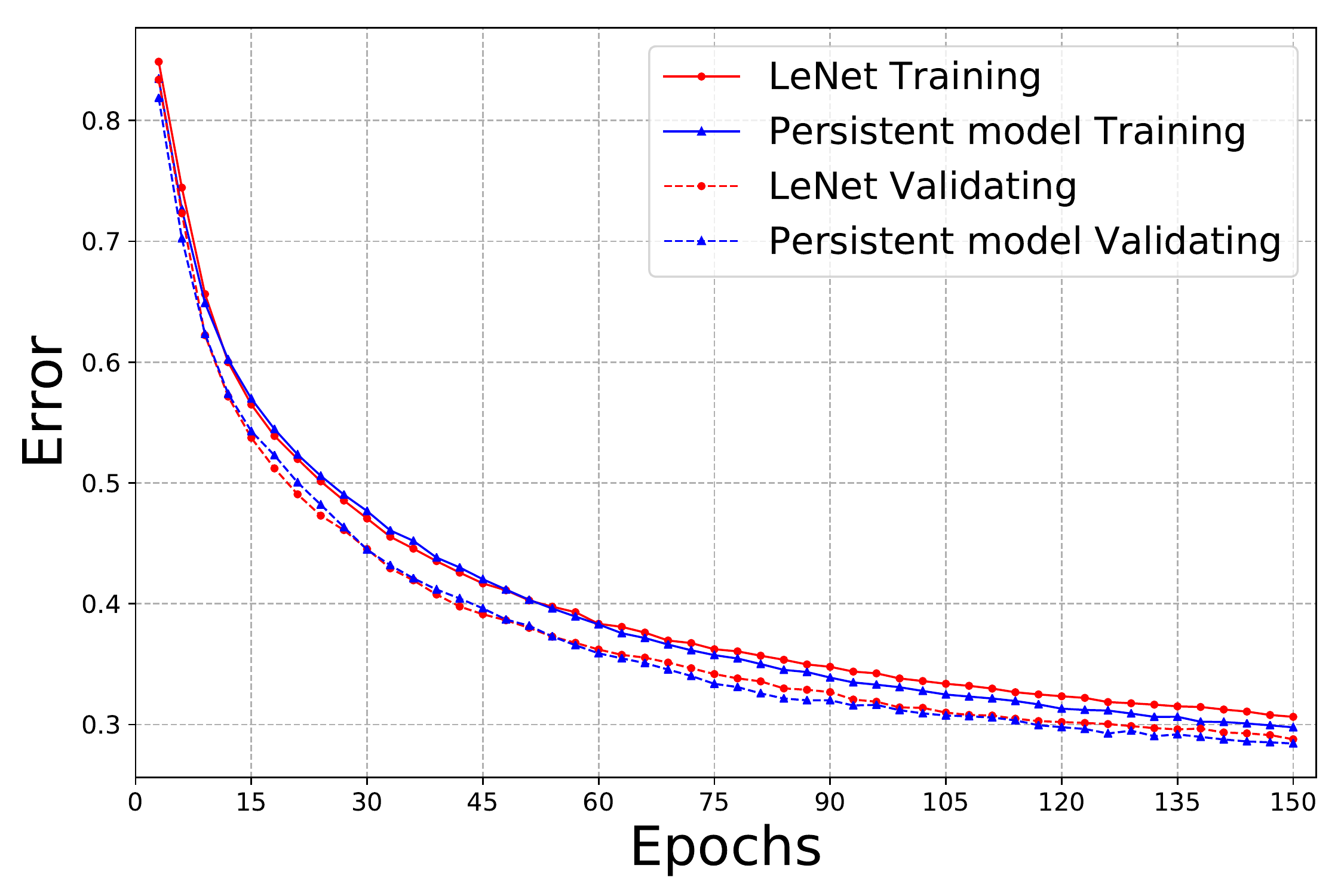} 
    \end{minipage}\hfill
    \begin{minipage}{0.5\textwidth}
        \centering
        \includegraphics[width=0.9\textwidth]{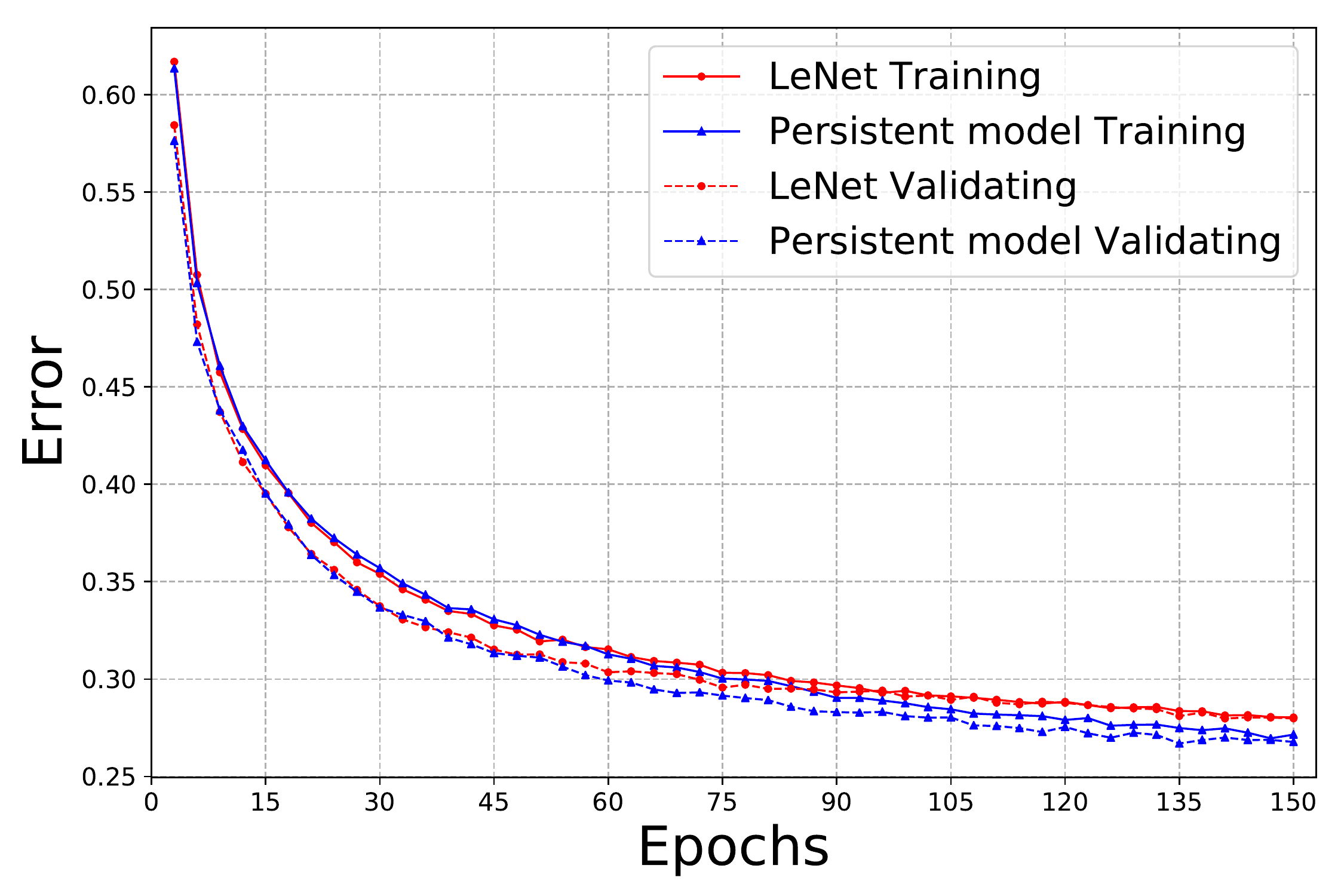} 
    \end{minipage}
    \caption{left: Validation and test error of  persistent iterations on CIFAR-10 (Initialization 1); right: Validation and test error of  persistent iterations on CIFAR-10 (Initialization 2).}
   \label{fig:lenet2}
\end{figure}

Figure~\ref{fig:lenet1} shows the persistent training results. The red and blue  solid horizontal lines show the validation and test error for the plain LeNet-5, the dashed lines denote the errors of persistent training. The plain model with initialization 1($\boldsymbol{\Theta}^{ini}_1$) has a validation accuracy of 70.74\% with test accuracy 69.94\%, while for initialization 2($\boldsymbol{ \Theta_{ini_2}}$) the validation accuracy is 71.58\% with test accuracy 71.48\%.  After persistent training on $\boldsymbol{\Theta_{ini_1}}$ 
for 19 iterations(shown in Figure~\ref{fig:lenet2}), the championship validation accuracy is boosted from 70.74\% to 71.64\% with test accuracy fromto 69.94\% to 71.78\%, surpassing the well-initialized $\boldsymbol{\Theta_{ini_2}}$ scenario. 
This suggests that persistent training can make up the gap between poorly-born and well-born neurons. 

Furthermore, we apply persistent training on well-born neurons. As shown in Figure~\ref{fig:lenet1}, persistent training also helps improve the model with $\boldsymbol{\Theta_{ini_1}}$(validation accuracy =  72.76\% and test accuracy = 73.22\%). 

Figure~\ref{fig:lenet2} shows the training curves corresponding to the championship persistent model with $\boldsymbol{\Theta_{ini_1}}$ and $\boldsymbol{\Theta_{ini_2}}$. 

\paragraph{Partial Persistent Model}
The full weight persistent model leverages all layer's previous converged weight during persistent training. This section introduces the partial persistent model and applies it to a variety of architectures. The partial persistent model only takes a random layer's parameters into persistent training. Let $l_s$ be a random layer. The loss for $n_{th}$ iteration is:
\begin{equation}
    \mathcal{L}_{n}({\theta}) = \mathcal{L}({\theta}) + \lambda   \sum_{k=0}^{n-1} \sum_{l=0}^m \frac{{|\boldsymbol{\Theta}^l_{k}}^T{{\theta}^l}| }{ || \boldsymbol{\Theta}^l_k ||^2} \delta{(l-l_s)}
\end{equation}

We apply partial persistent training on LeNet-5(batch size=256), AlexNet and ResNet-18(batch size=128) for 150, 150 and 350 epochs using Adam optimizer \cite{kingma2014adam} with learning rate 0.001. The persistent hyperparameter $\lambda$ is set as 0.001.   Default weights initialization  scheme in PyTorch  \cite{paszke2019pytorch} is used for all the layers. 

\begin{figure}[h]
    \centering
    \begin{minipage}{0.5\textwidth}
        \centering
        \includegraphics[width=0.95\textwidth]{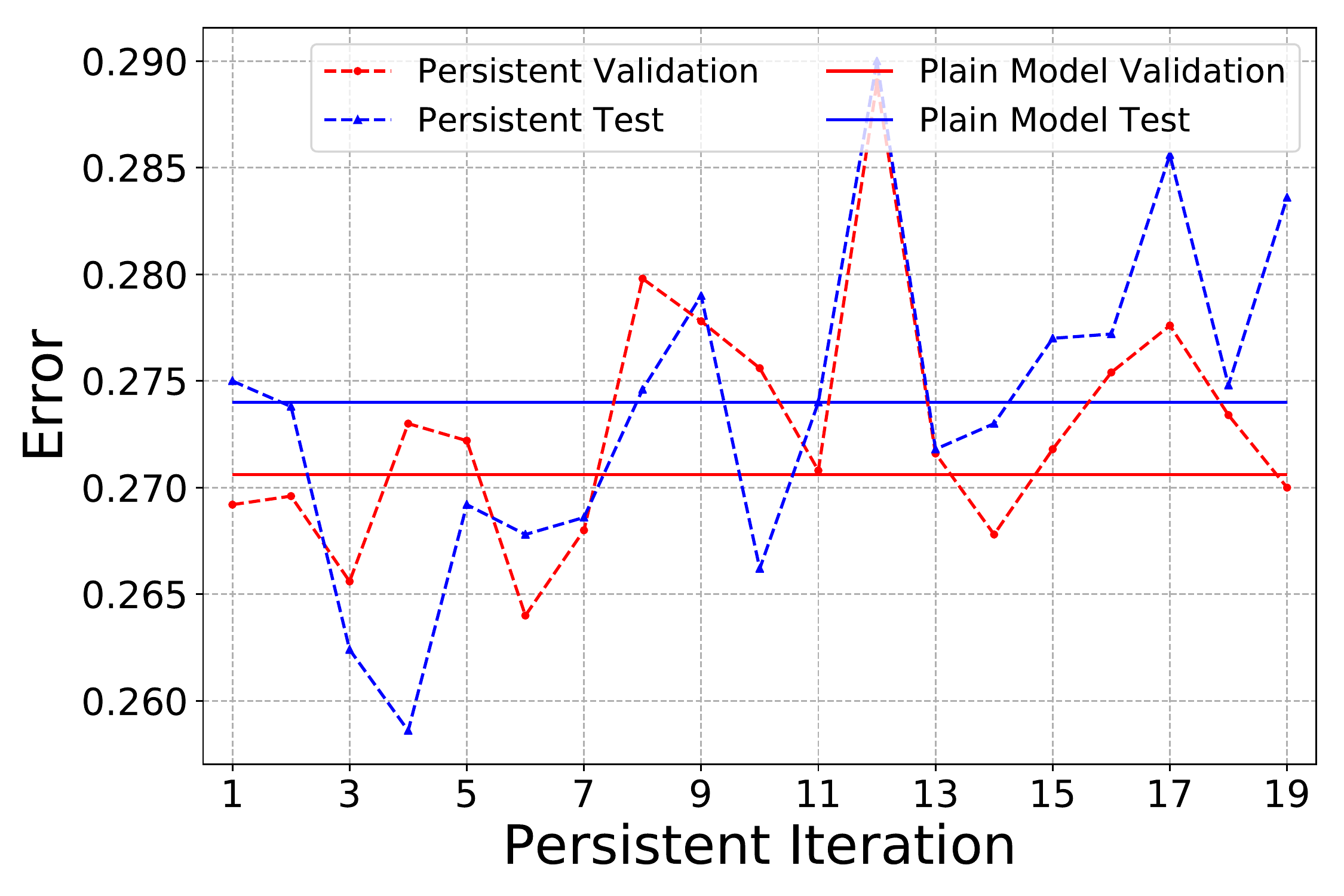} 
    \end{minipage}\hfill
    \begin{minipage}{0.5\textwidth}
        \centering
        \includegraphics[width=0.95\textwidth]{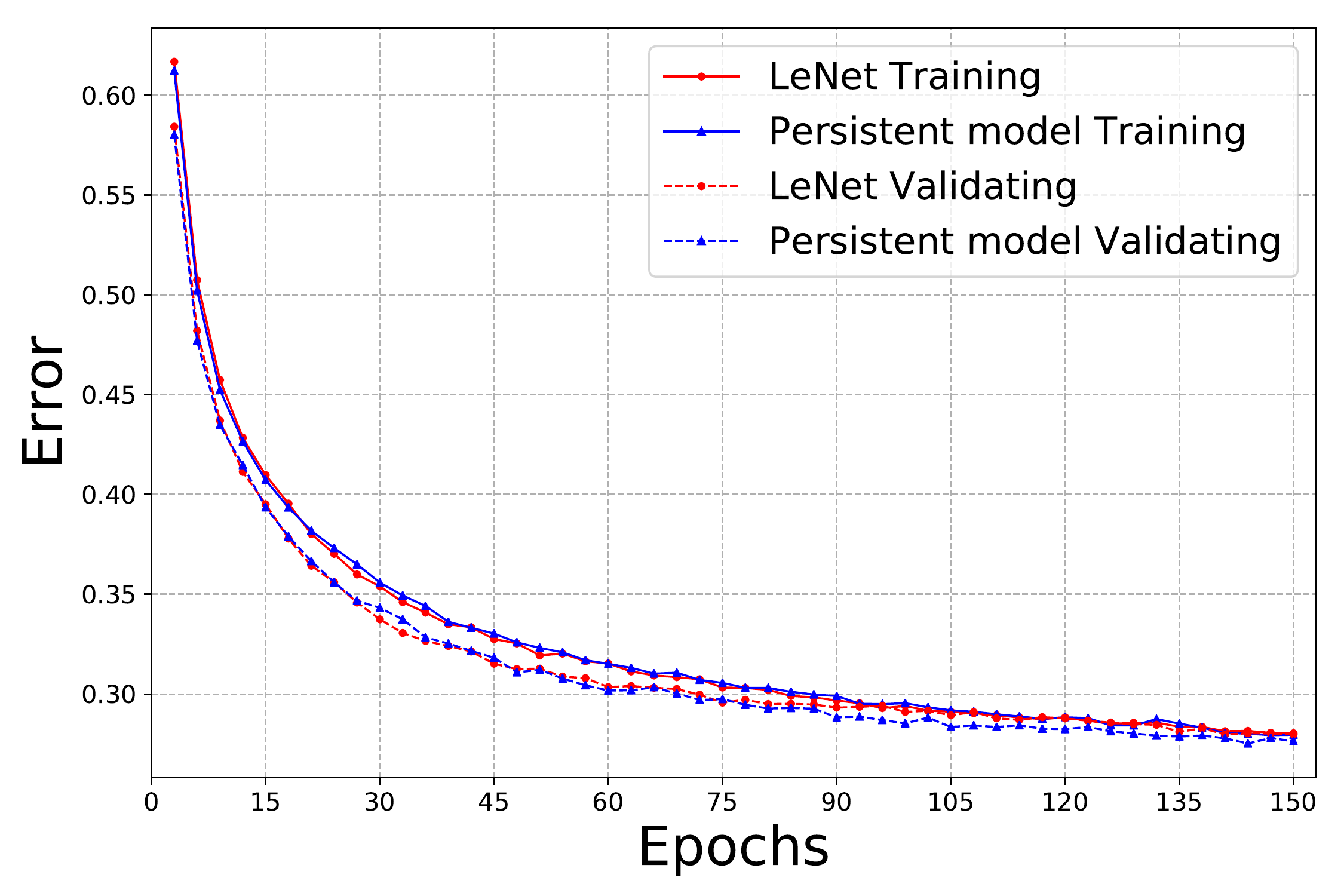} 
    \end{minipage}
    \caption{left: Validation and test error of different persistent iteration on CIFAR-10 (LeNet-5); right: Training and validating history (LeNet-5).}
   \label{fig:partialLenet}
\end{figure}

\begin{figure}[h]
    \centering
    \begin{minipage}{0.5\textwidth}
        \centering        \includegraphics[width=0.95\textwidth]{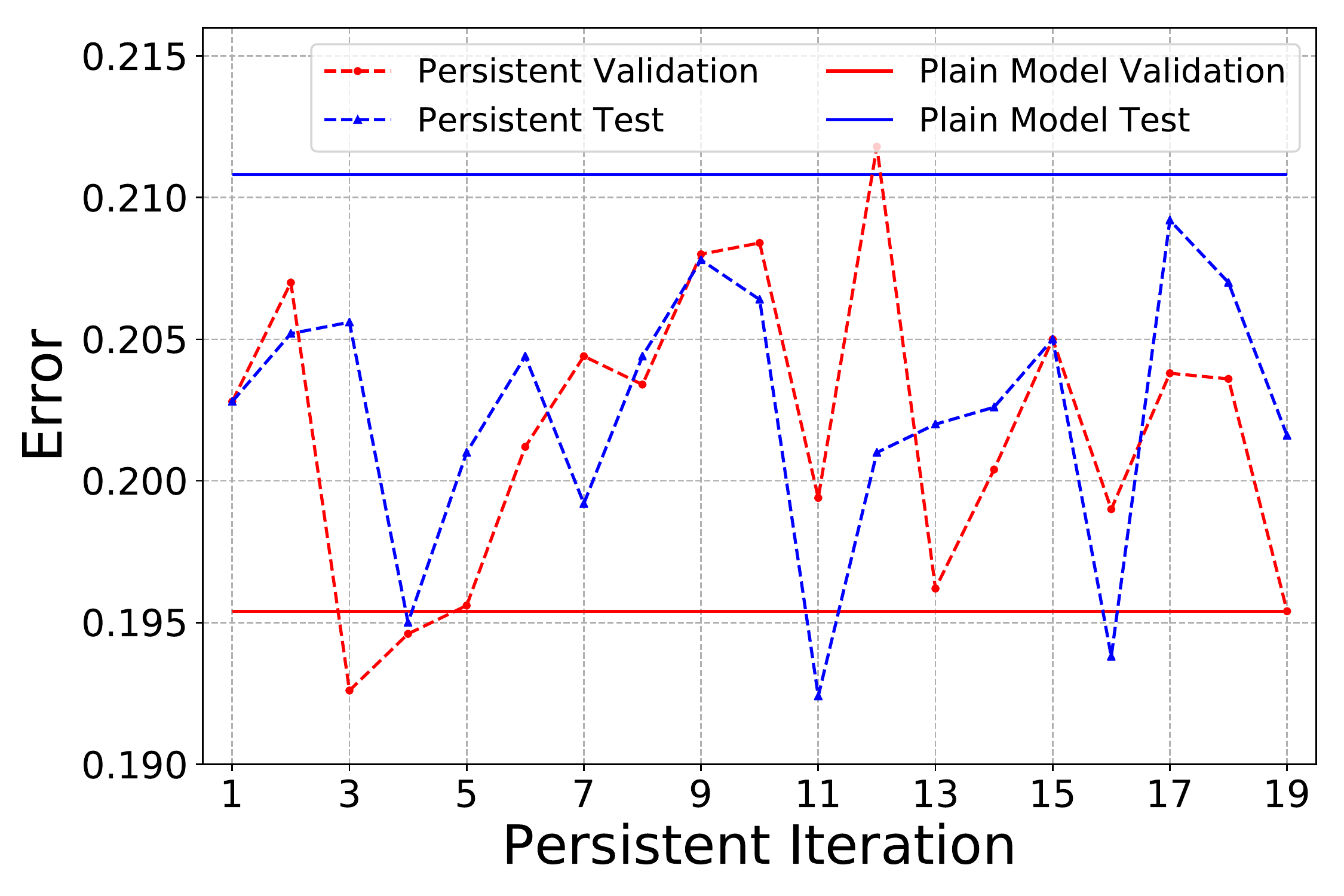} 
    \end{minipage}\hfill
    \begin{minipage}{0.5\textwidth}
        \centering
        \includegraphics[width=0.95\textwidth]{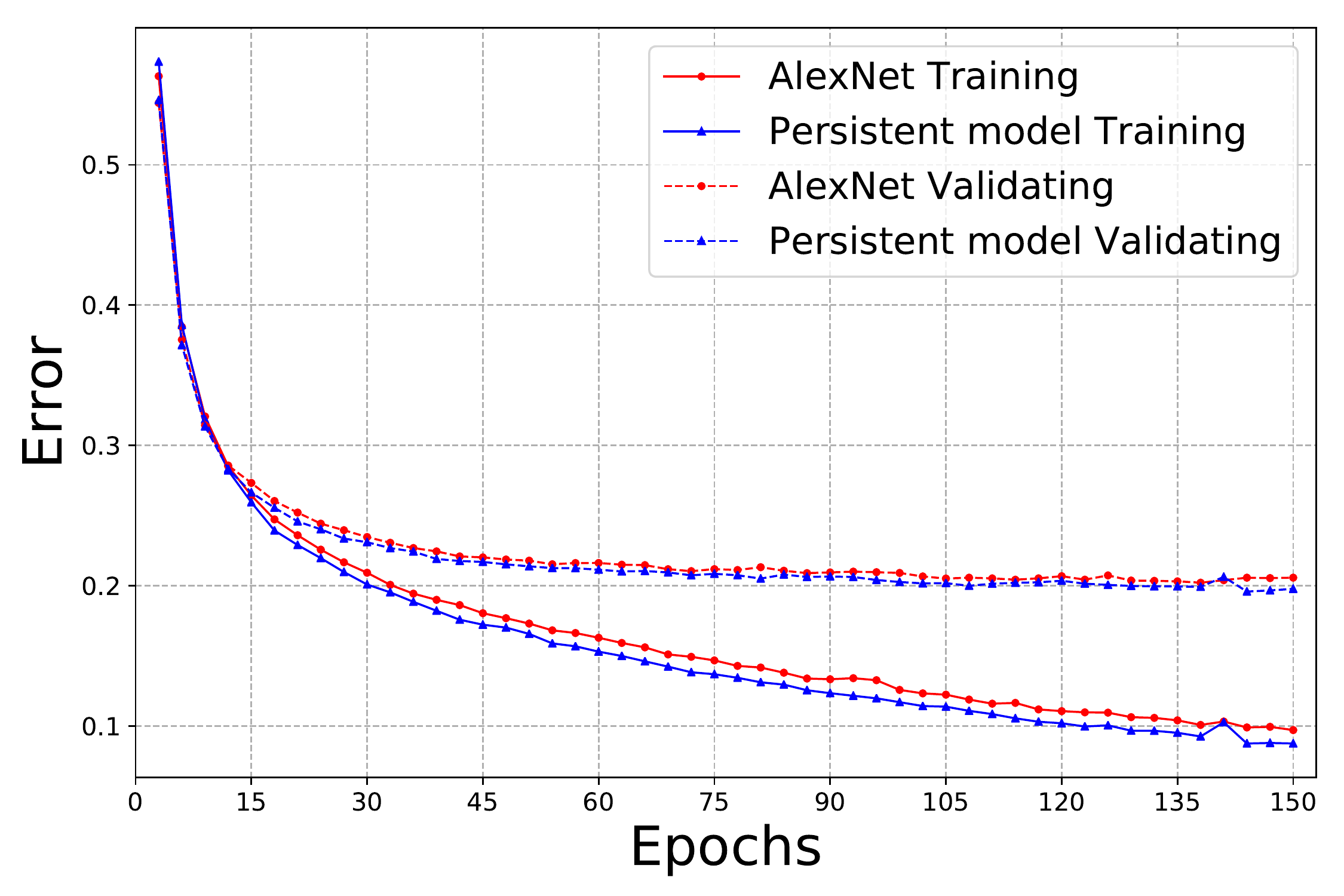} 
    \end{minipage}
    \caption{left: Validation and test error of different persistent iteration on CIFAR-10 (AlexNet); right: Training and validating history (AlexNet).}
   \label{fig:partialAlex}
\end{figure}
\begin{figure}[h]
    \centering
    \begin{minipage}{0.5\textwidth}
        \centering        \includegraphics[width=0.95\textwidth]{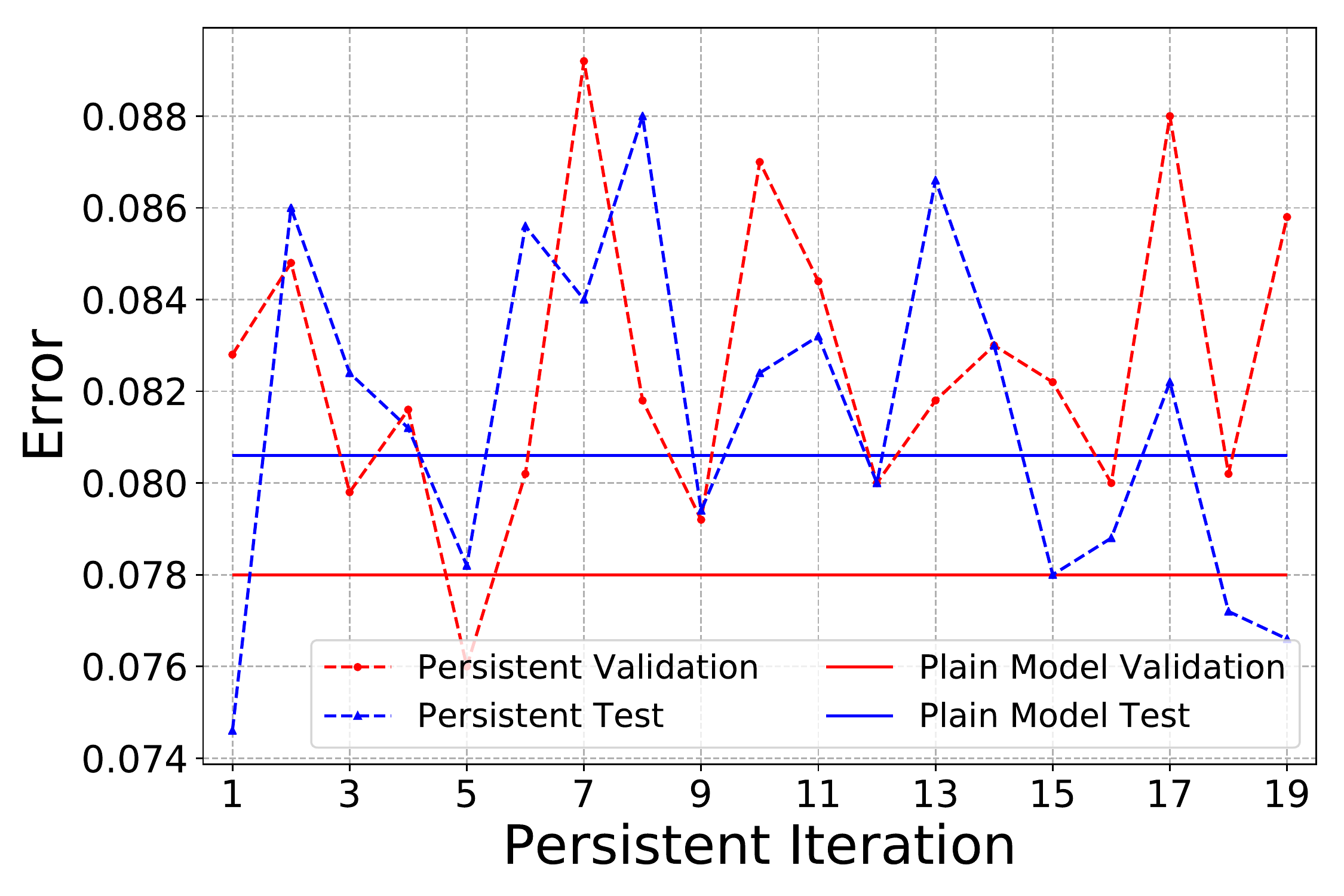} 
    \end{minipage}\hfill
    \begin{minipage}{0.5\textwidth}
        \centering
       \includegraphics[width=0.95\textwidth]{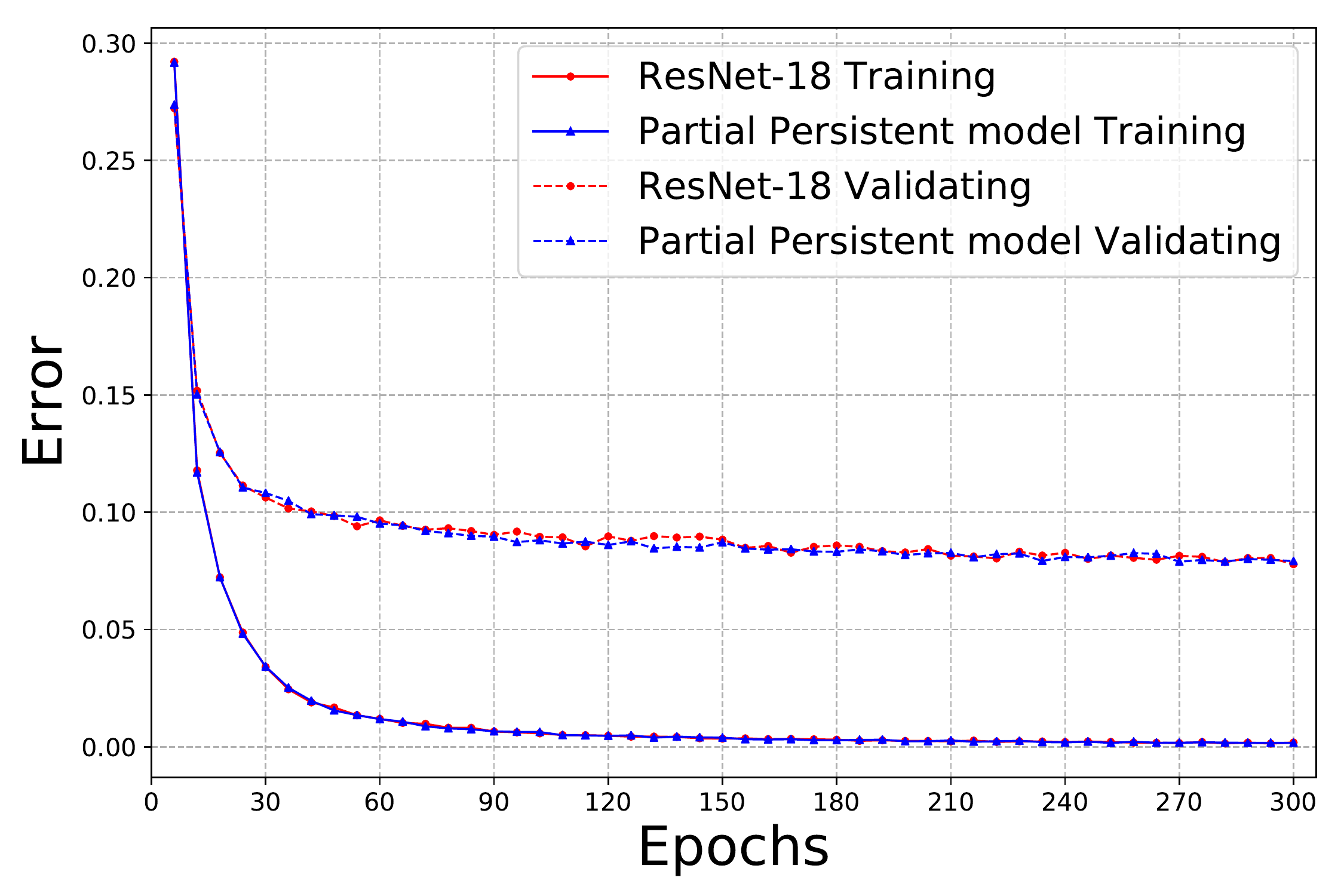} 
    \end{minipage}
    \caption{left: Validation and test error of different persistent iteration on CIFAR-10 (ResNet-18); right: Training and validating history (ResNet-18).}
   \label{fig:partialResNet}

\end{figure}
Figure~\ref{fig:partialLenet}, \ref{fig:partialAlex} and \ref{fig:partialResNet} show the persistent training results on the three networks and championship models' training curves. Our observations suggest that persistent training helps improve the models' accuracy. For LeNet-5, the championship persistent model achieves validation accuracy 73.60\% and test accuracy 73.22\% while for plain model the accuracies are 72.94\% and 72.60\%; persistent training boosts the validation accuracy from 80.46\% to 80.74\% and test accuracy from 78.92\% to 79.44\% on AlexNet; the validation accuracy increases from 92.20\% to 92.40\% and test accuracy from 91.94\% to 92.18\% on ResNet-18. 

Our results also suggest that ResNet, though considered with smooth loss landscape geometry where non-convexities should not be problematic \cite{draxler2018essentially} \cite{li2018visualizing}, can still converge to sub-optimal regions. 
The persistent training drives the weights leaving the sub-optimal solutions and arriving at new locations with improved validation accuracy. 
Our empirical results suggest that ResNet's loss landscape can still have different optima.
\paragraph{Gradients Update in Full and Partial Model}
The gradients for $n_{th}$ full persistent model w.r.t. $\theta_l$ are:
\begin{equation}\label{eq:gradgln1}
g^l_n = \nabla_{\theta_l}   \mathcal{L}({\theta}) + \lambda  \sum_{k=0}^{n-1} \frac{sign({\boldsymbol{\Theta}^l_{k}}^T{{\theta}^l}) }{ || \boldsymbol{\Theta}^l_k ||^2} \boldsymbol{\Theta}^l_{k} 
\end{equation}
Recall $\boldsymbol{\Theta}^l_{k}$ represents the flattened converged solution at layer $l$ after $k_{th}$ persistent training and $\theta^l$ is the parameters to be optimized at layer $l$. Let the $\mathbb{E}$ $\big[ \sum_{k=0}^{n-1} \frac{sign({\boldsymbol{\Theta}^l_{k}}^T{{\theta}^l}) }{ || \boldsymbol{\Theta}^l_k ||^2} \boldsymbol{\Theta}^l_{k} \big]$ $= $ $\boldsymbol{C}^l_n$, $\boldsymbol{C}^l_n$ is bounded as $sign({\boldsymbol{\Theta}^l_{k}}^T{{\theta}^l}) \in \{ -1,1 \}$ and $\boldsymbol{\Theta}^l_k$ is a set of constants. So $g^l_n$ is:
\begin{equation}\label{eq:gradgln2}
g^l_n = \nabla_{\theta_l}   \mathcal{L}({\theta})+ \lambda \boldsymbol{C}^l_n  + \lambda  d^l_n(\theta^l) 
\end{equation}
where $ d^l_n(\theta^l)$ $=$ $\sum_{k=0}^{n-1} \frac{sign({\boldsymbol{\Theta}^l_{k}}^T{{\theta}^l}) }{ || \boldsymbol{\Theta}^l_k ||^2} \boldsymbol{\Theta}^l_{k} - \boldsymbol{C}^l_n$, $\mathbb{E} \left[ d^l_n{(\theta^l})\right] = 0$.
Thus, $g^l_n$ is the sum of the plain model's derivative $\nabla_{\theta_l} \mathcal{L}({\theta})$, bounded constant $\lambda \boldsymbol{C}^l_n$ and zero-mean noise term $\lambda d^l_n(\theta^l)$. The full gradients: $\nabla_{\theta}   \mathcal{L}_n = \left[g^1_n\Vert g^2_n \Vert...\Vert g^m_n\right]$ is the concatenation of $g^l_n$. Thus, the gradient descent during $n_{th}$ persistent training can be written:
\begin{equation}
    \theta_{t+1} = \theta_t - \eta g_n 
\end{equation}
where
\begin{align*}
    & g_n = \nabla_{\theta}  \mathcal{L}({\theta}) +\\
    & \lambda \big\{ \underbrace{ \left[\boldsymbol{C}^1_n\Vert \boldsymbol{C}^2_n \Vert...\Vert \boldsymbol{C}^m_n\right] }_{bias} +\underbrace{\left[d^1_n\Vert d^2_n \Vert...\Vert d^m_n\right] }_{zero-mean~noise} \big\}    
\end{align*}
So persistent neurons can be regarded as a biased stochastic gradient method, where individual updates are corrupted by biased error terms. 
The convergence of biased stochastic gradient method have been discussed by previous researchers \cite{chen2018stochastic}\cite{ajalloeian2020analysis}\cite{hu2020biased}. Biased stochastic gradient methods can in general converge to a neighborhood of the solution,  but the optimum still can be reached under special cases, we refer to \cite{ajalloeian2020analysis} for the detailed analysis of convergence results.

\section{Neuron Dynamics in Persistent Training}
\paragraph{Do we find the better solution using persistent training?}
As previous mentioned,  error loss function presents \textbf{few extremely} wide flat minima (WFM) which coexist with narrower minima and critical points \cite{baldassi2020shaping}. This suggests the minima are different. Do we find the better minima using persistent training? 
\begin{figure}[t]
    \centering
    \begin{minipage}{0.5\textwidth}
        \centering
        \includegraphics[width=1.0\textwidth]{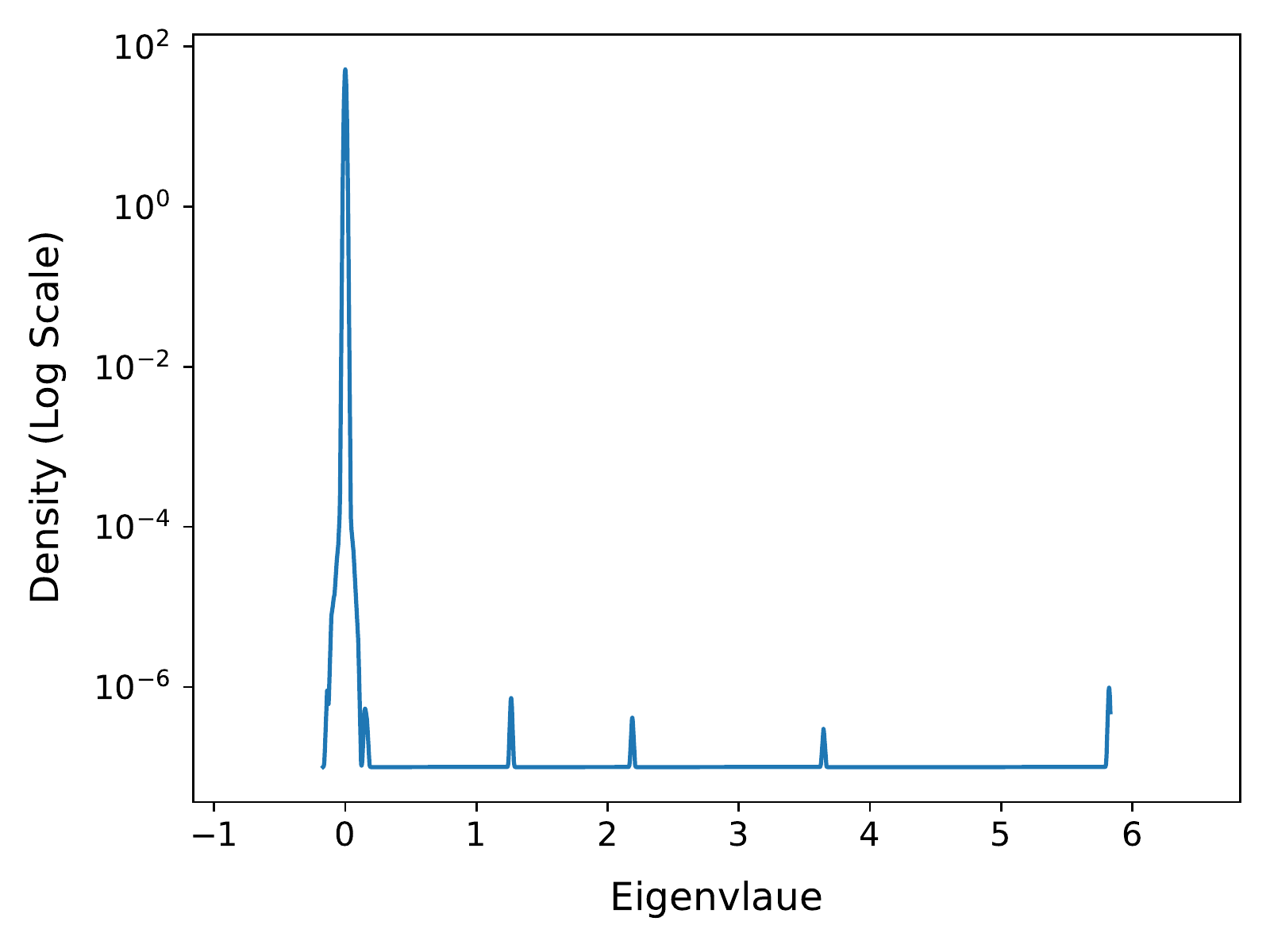} 
    \end{minipage}\hfill
    \begin{minipage}{0.5\textwidth}
        \centering
        \includegraphics[width=1.0\textwidth]{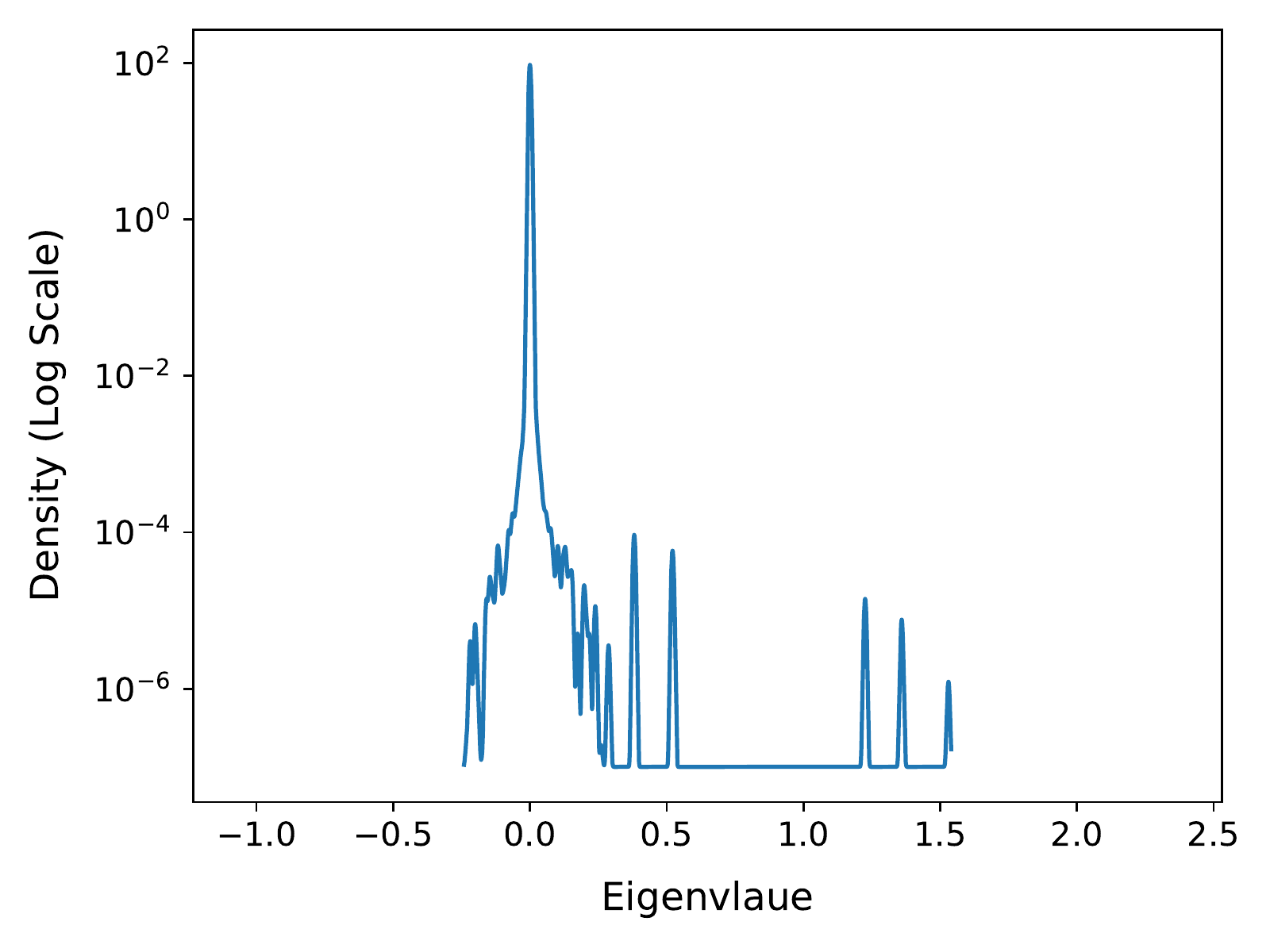} 
    \end{minipage}
    \caption{Spectral densities of ResNet-18. Left:  vanilla ResNet-18; right: Championship persistent training model. Persistent training further shortened the distance between the largest eigenvalue of the Hessian and the bulk.}
    \label{fig:Spectraldensities}
\end{figure}
\begin{figure}[H]
\centering
\centerline{\includegraphics[width=0.5\textwidth]{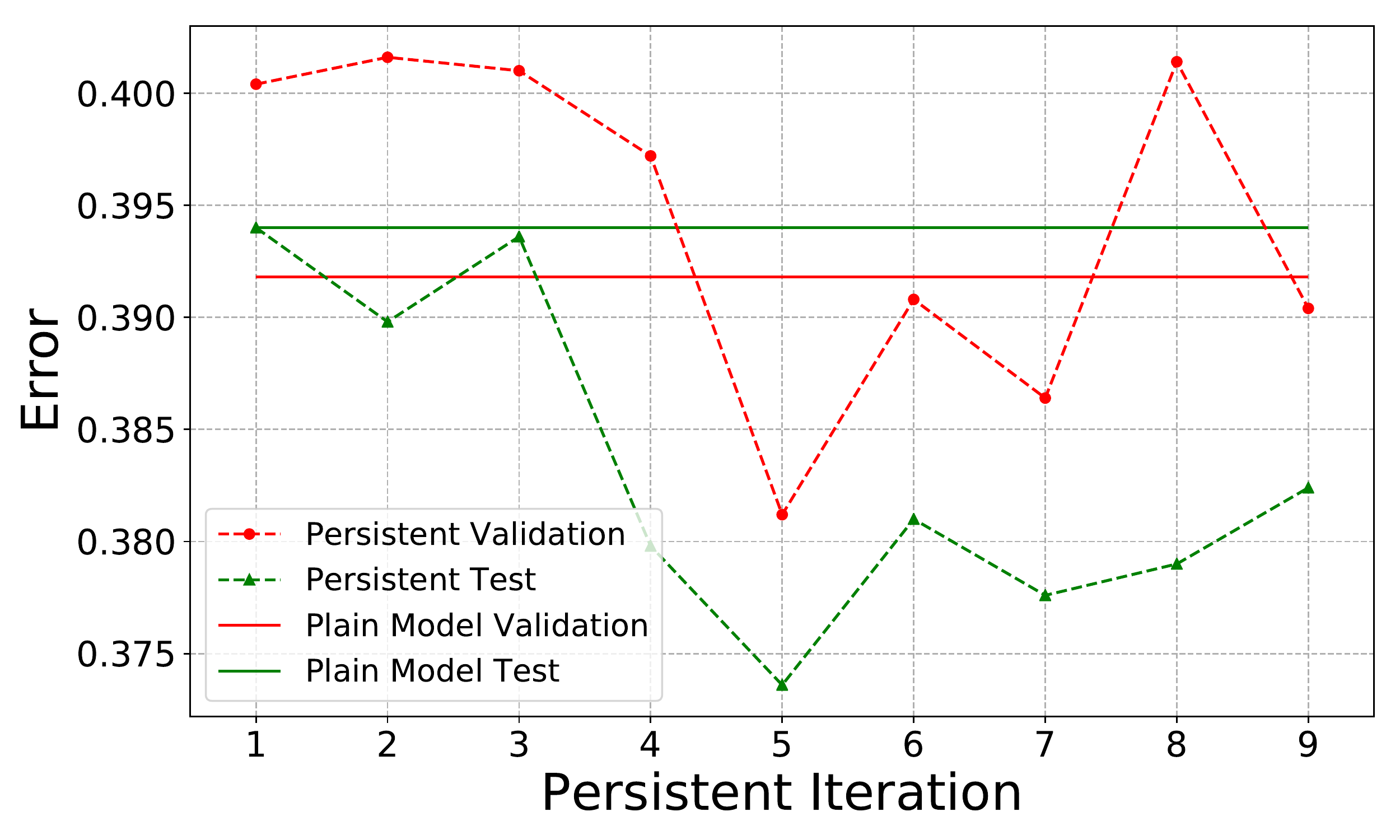}}
\caption{Validation and test error of different persistent iterations using a 7 layer NN.}
\label{fig:satuat}
\end{figure}  
For Partial Lenet and Partial Alex, the validation curve clearly show that the persistent training leads to lower validation error, as shown in the training history in Figure~\ref{fig:partialLenet} and \ref{fig:partialAlex}. For ResNet-18, the gap between plain model and persistent model in Figure~\ref{fig:partialResNet} is hard to distinguish, here we compare the vanilla solution and the championship persistent solution via Hessian eigenvalue density.  As suggested by \cite{ghorbani2019investigation}, outliers in Hessian spectrum hurt the optimization and batch normalization leads to a better solution with suppressed outliers in  Hessian spectrum. Our results are shown in Figure~\ref{fig:Spectraldensities}, where we compare the persistent championship's Hessian spectrum with the plain model's spectrum, we found that the championship persistent model's spectrum has fewer  outliers than the plain model, suggesting a better minima\footnote{We follow the original setting in \cite{he2016deep}, the vanilla ResNet-18  already adopts batch normalization in the model.}.





To further study how persistent neurons helps training, we investigate the saturation of the network. Saturation
is an important feature that can be used as a descriptor of the training process, as well as understand the behaviour of the network itself \cite{kolbusz2018neural} \cite{rakitianskaia2015measuring}. A better trajectory should have less saturated neurons during training, thus beneficial for the learning tasks \cite{glorot2010understanding}\cite{he2015delving}. 
\begin{figure}[htb!]
    \centering
    \begin{minipage}{0.48\textwidth}
        \centering
        \includegraphics[width=1.0\textwidth]{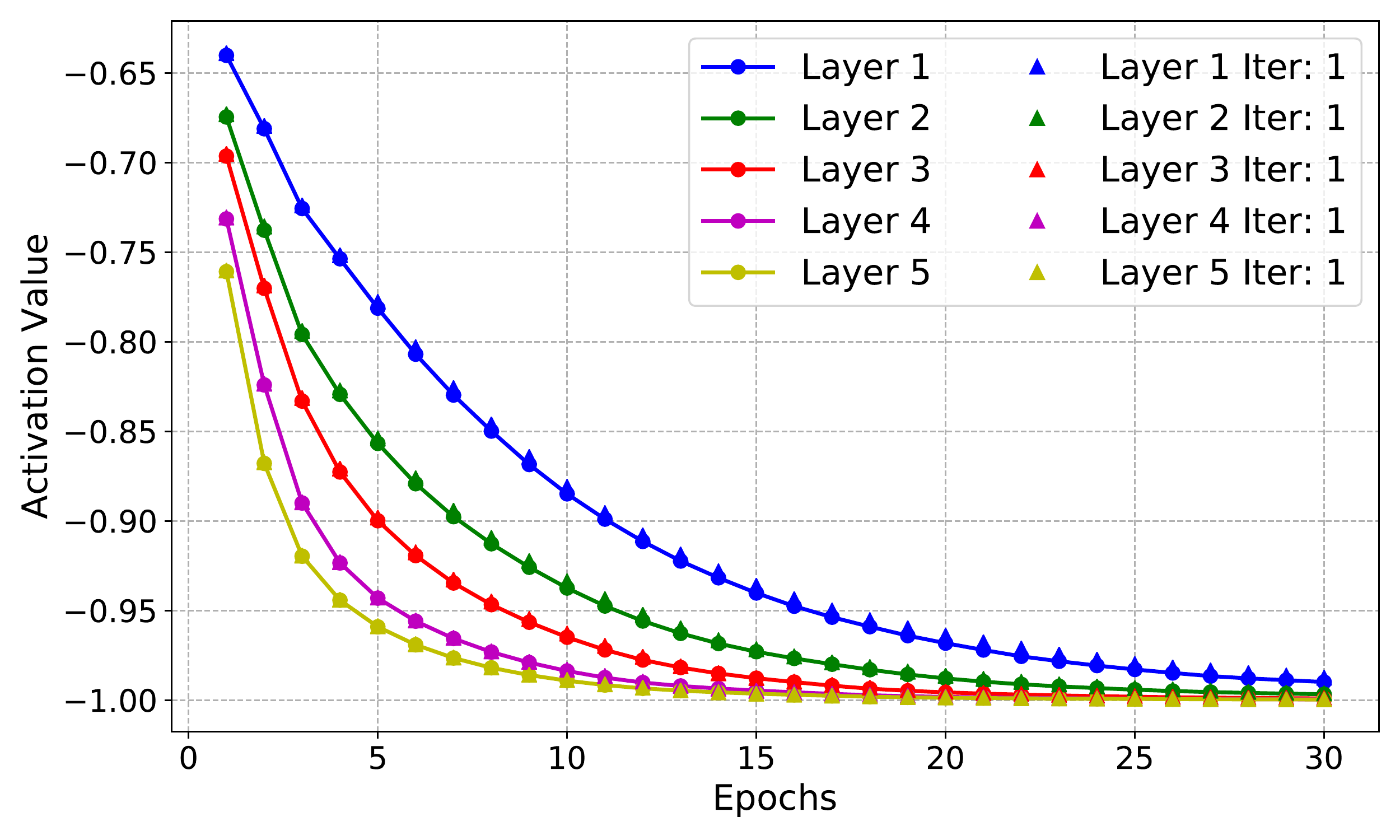} 
        \caption*{$1_{st}$ persistent training}
    \end{minipage}\hfill
    \begin{minipage}{0.48\textwidth}
        \centering
        \includegraphics[width=1.0\textwidth]{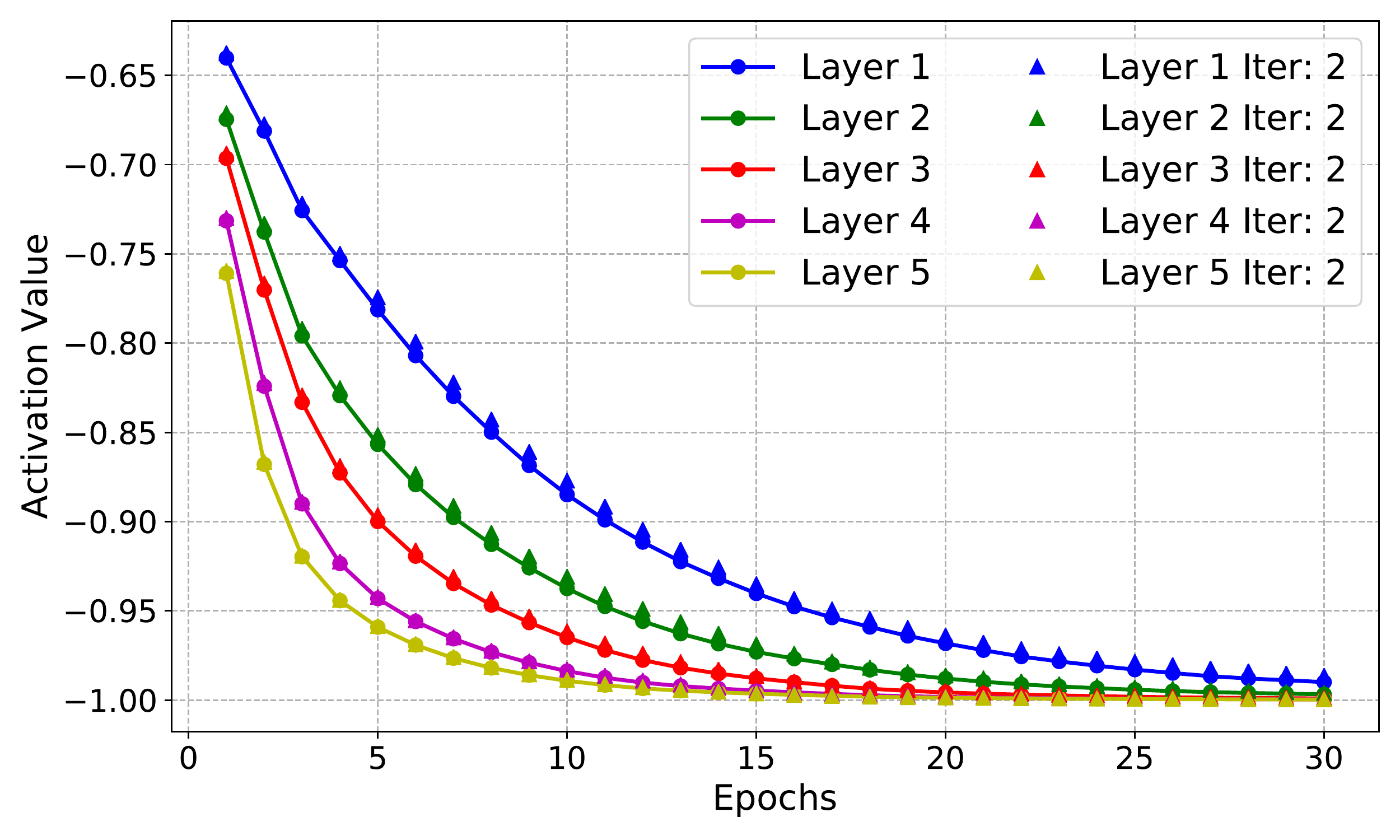} 
        \caption*{$2_{nd}$ persistent training}
    \end{minipage}\hfill
        \begin{minipage}{0.48\textwidth}
        \centering
        \includegraphics[width=1.0\textwidth]{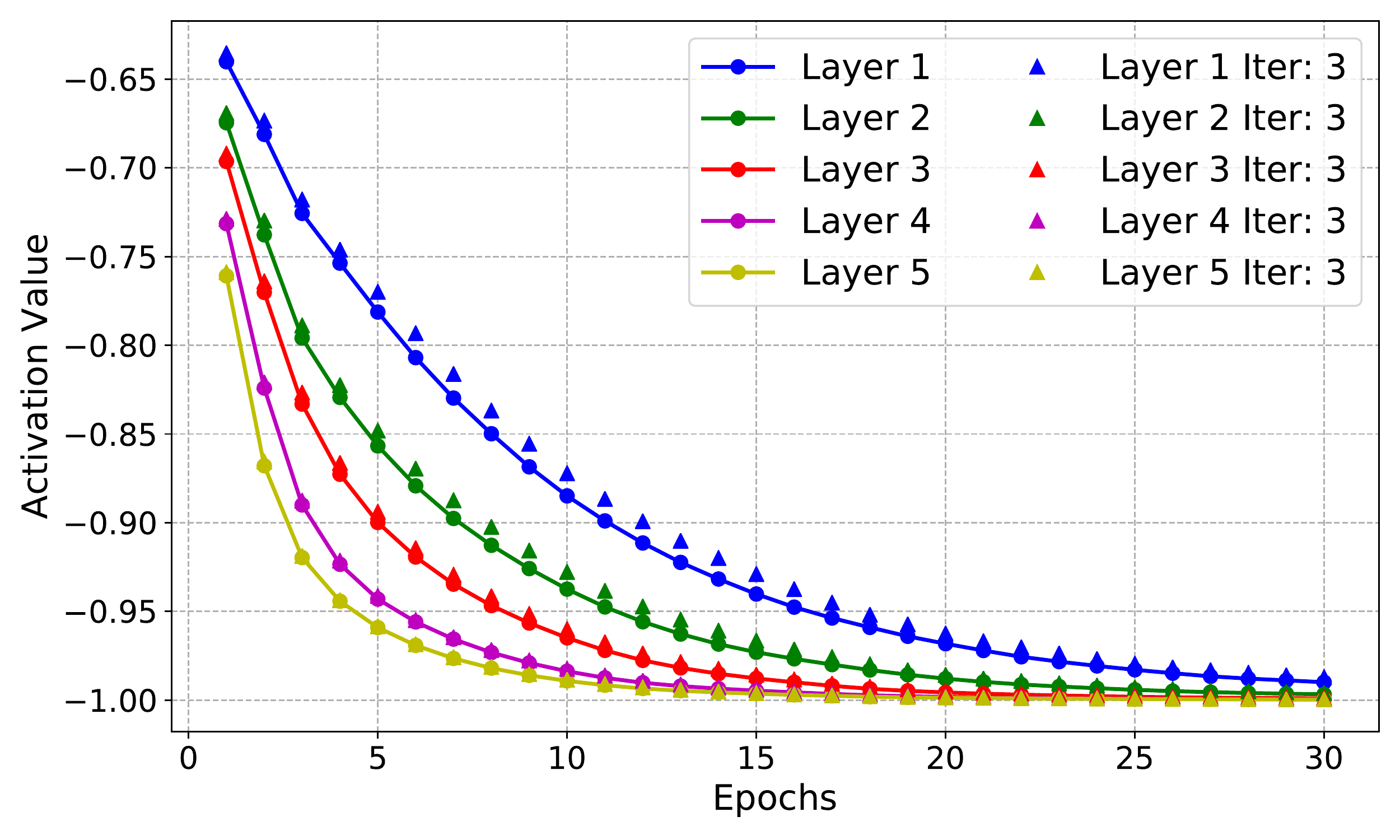} 
        \caption*{$3_{rd}$ persistent training}
    \end{minipage}
    \begin{minipage}{0.48\textwidth}
        \centering
        \includegraphics[width=1.0\textwidth]{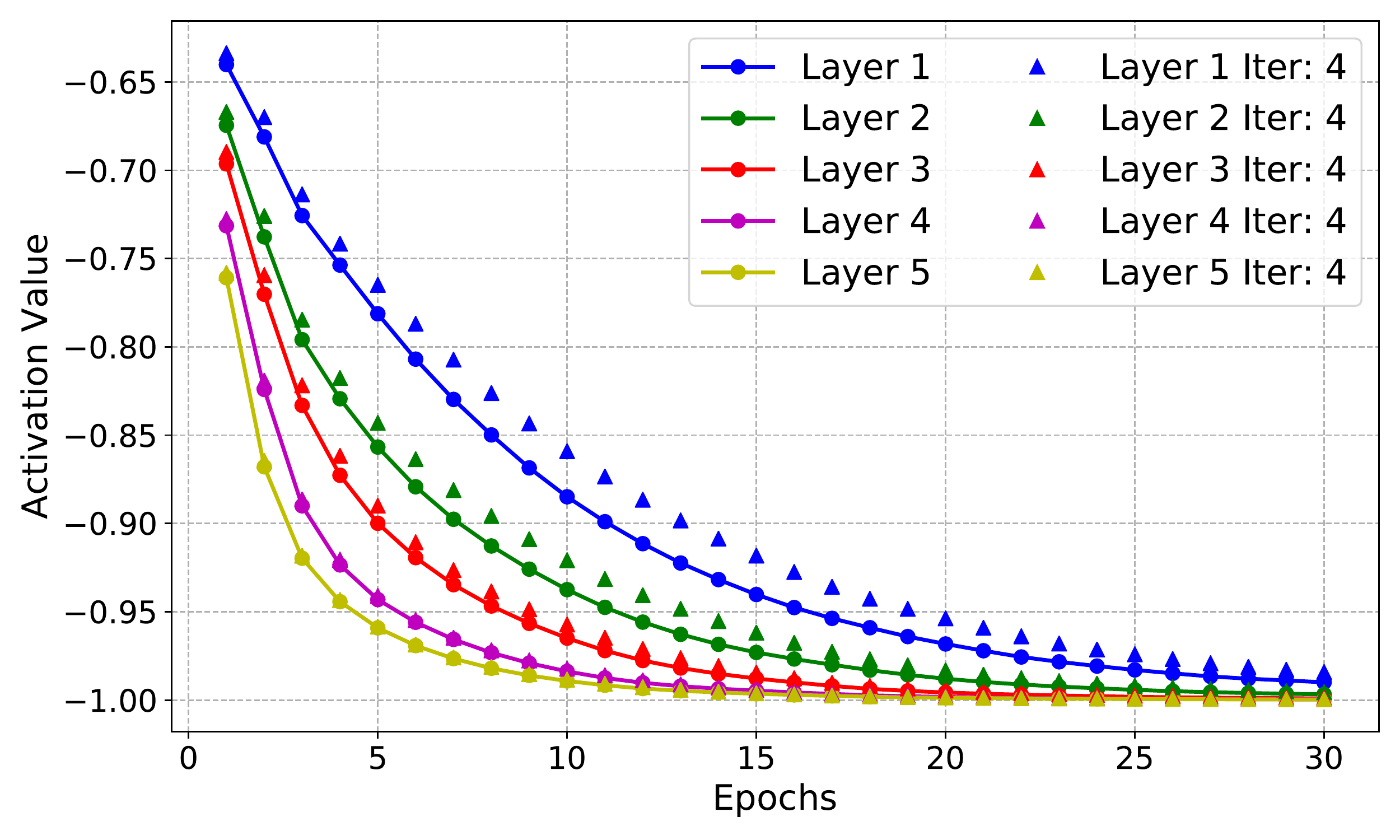} 
        \caption*{$4_{th}$ persistent training}
    \end{minipage}\hfill
        \begin{minipage}{0.48\textwidth}
        \centering
        \includegraphics[width=1.0\textwidth]{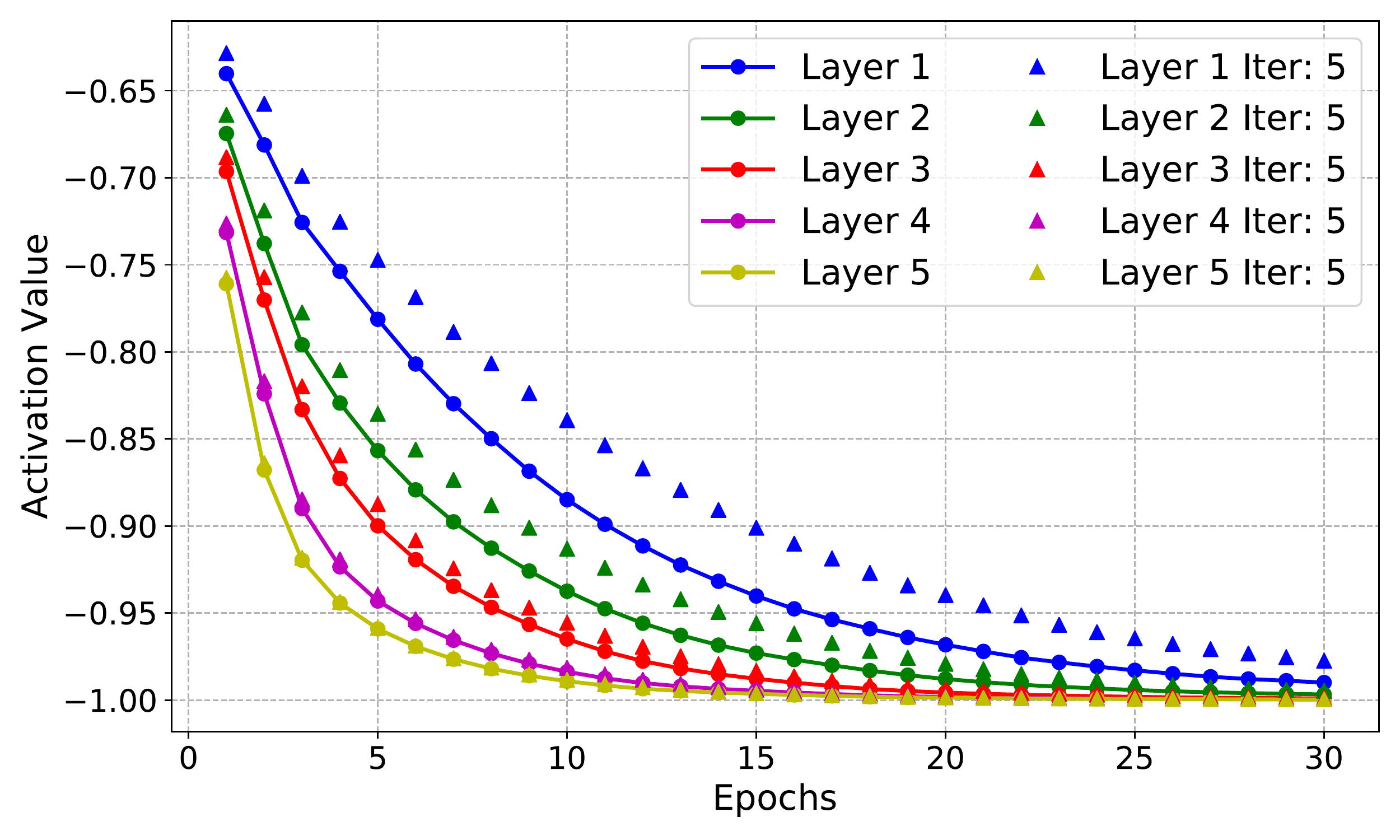} 
        \caption*{$5_{th}$ persistent training}
    \end{minipage}\hfill
    \begin{minipage}{0.48\textwidth}
        \centering
        \includegraphics[width=1.0\textwidth]{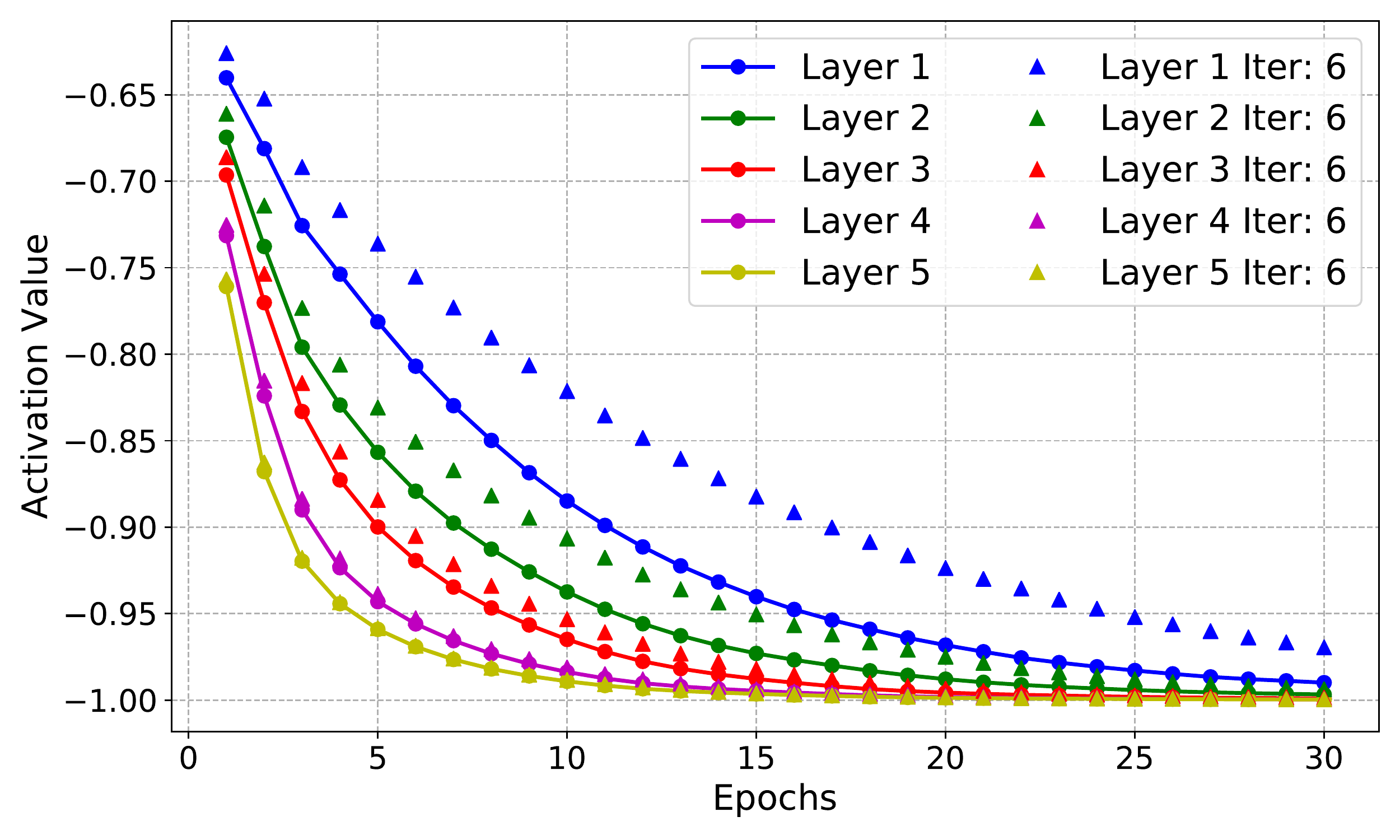} 
        \caption*{$6_{th}$ persistent training}
    \end{minipage}
\caption{98 percentiles of the distribution of the activation values for the hyperbolic tangent networks in the course of learning(training for 30 epochs). $5_{th}$ is the championship model with lowest validation error. The solid line is the plain model's results. }
\label{fig:satur}
\end{figure} 
Here we employ a 7 layer NN with Tanh activation for studying the saturation behaviour during partial persistent training. The CIFAR-10 dataset uses the same random split for test and validation as previous models.
Figure~\ref{fig:satuat} shows the persistent training history where the $5_{th}$ persistent training corresponds to the championship model.

Figure~\ref{fig:satur} shows the saturation behaviour during persistent training with hyperbolic tangent function as activation for CIFAR-10 classification with Xavier initialization  \cite{glorot2010understanding}. The neurons become less saturated as the persistent training iteration increases. During our persistent training, the $5_{th}$ iteration corresponds to the best performance. We note that the performance does not show monotonic behaviour with respect to persistent iterations, suggesting that a less saturated model does not always imply better accuracy.


\begin{figure}[H]
    \centering
    \begin{minipage}{0.5\textwidth}
        \centering
        \includegraphics[width=0.9\textwidth]{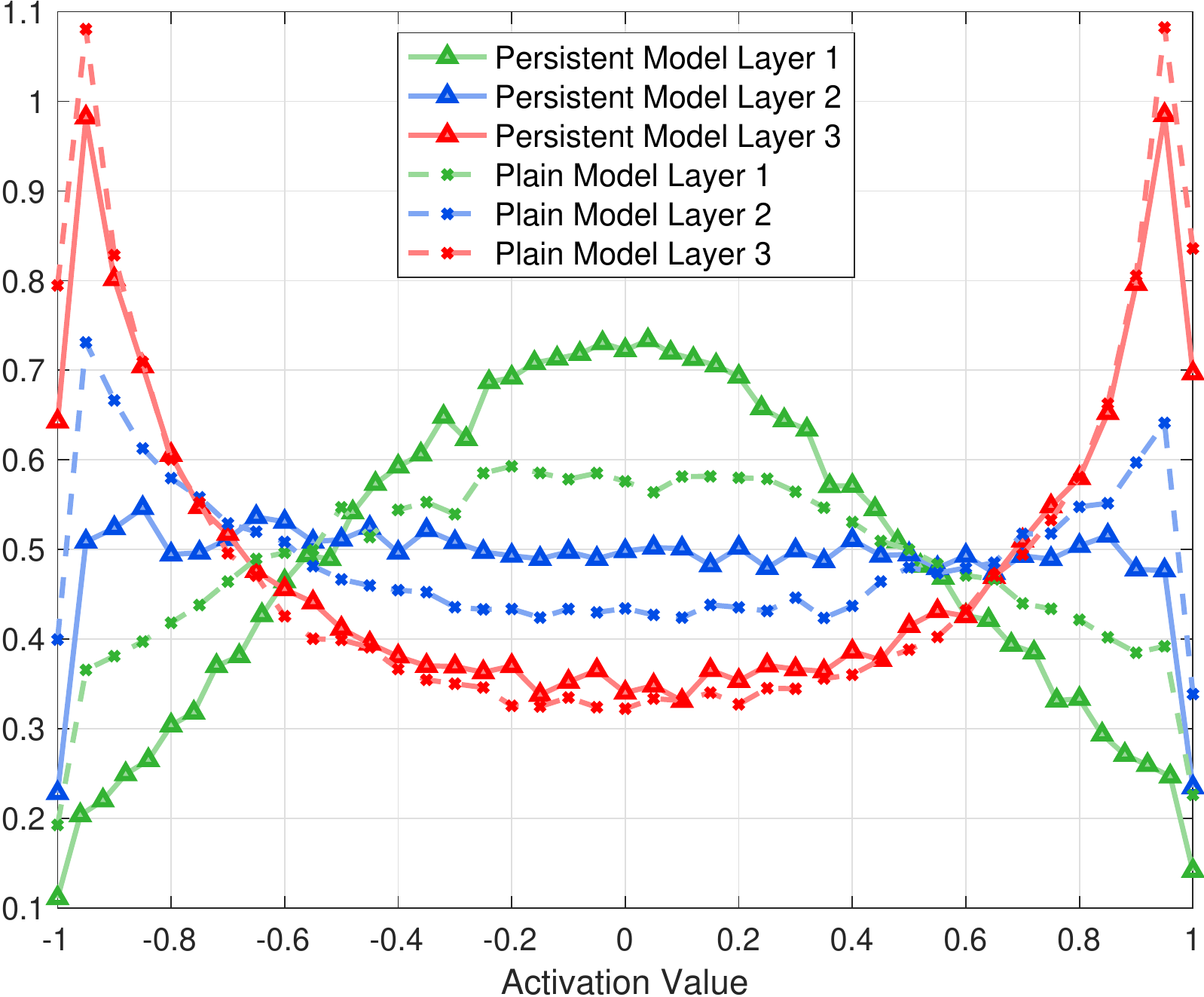} 
    \end{minipage}\hfill
    \begin{minipage}{0.5\textwidth}
        \centering
        \includegraphics[width=0.9\textwidth]{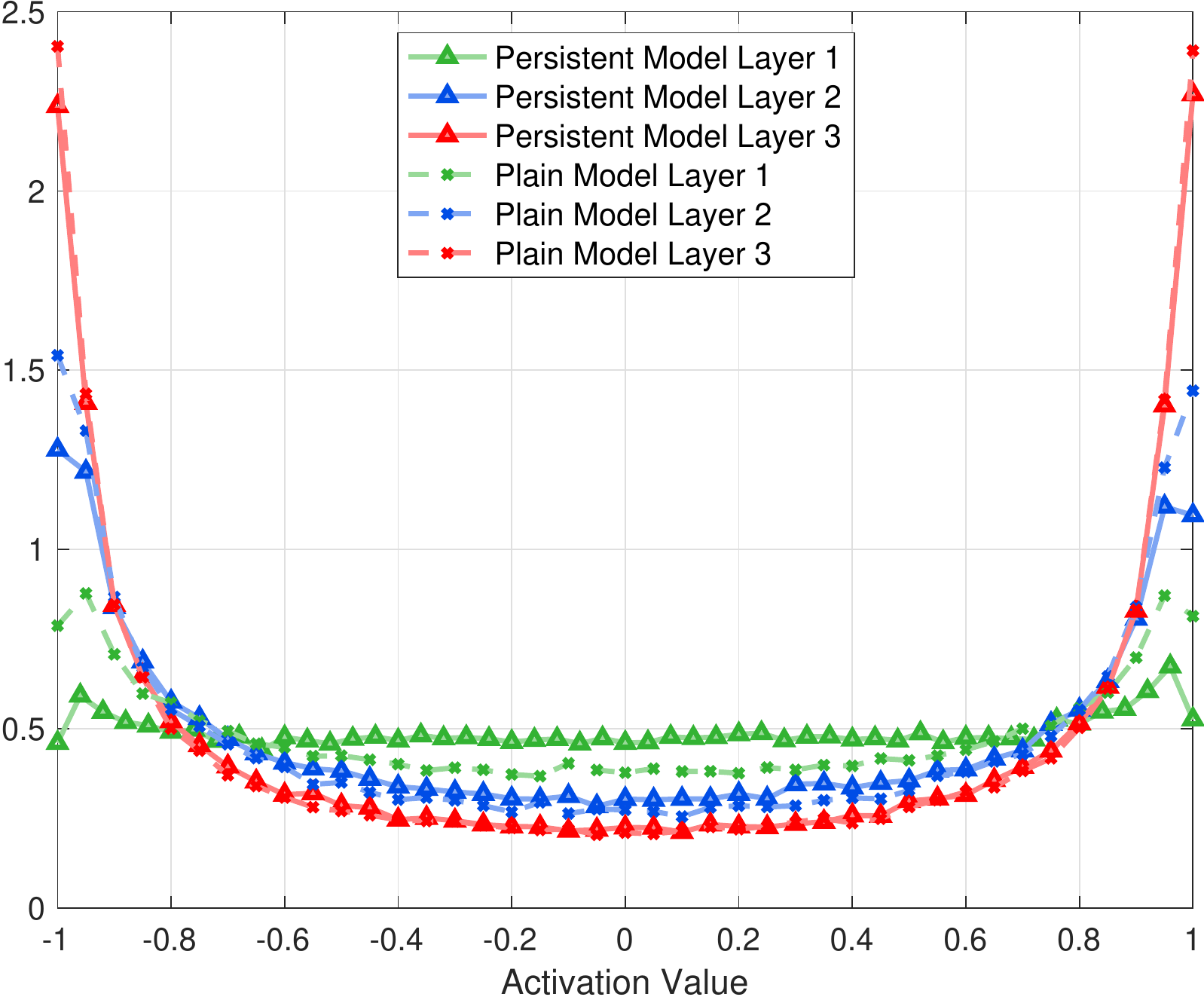} 
    \end{minipage}
    \caption{Activation values normalized histogram at $15_{th}$(left) and $30_{th}$(right) epoch.}
    \label{fig:epoch30}
\end{figure}
Figure~\ref{fig:epoch30} shows the activation values normalized histogram of the plain model and persistent championship model. The saturated plain model's activations distribute mostly at the extremes(asymptotes -1 and 1). The persistent training mitigates the saturation and re-distributes more weights on the linear or near-linear regions where the gradients can flow well.  As the number of persistent iteration increases, the parameters are repulsed from all previous converged minima/saddle points. These repulsion forces the model to explore different landscapes, resulting in a different path and less saturated behaviour. In persistent training, the championship model evolves within a certain saturation level, solely eliminating the saturation (increasing the persistent training iteration) does not boost the performance and can even weaken the model.

\section{Conclusion}
In this paper, we propose persistent neurons.
Using information from the previous training trajectories, persistent neurons drives the model to converge to new parameters under the same initialization.
Persistent neurons can be regarded as a biased stochastic gradient method.
We show that the standardized initialization methods, which solely utilize and analyze the start of the optimization trajectories, can fail on certain data distribution and persistent training helps overcome the problem and generalizes better.
This is achieved by incorporating additional information from the end of the trajectories, which is typically not leveraged in previous research.
We also show that persistent training achieves gains in performance in both well-initialized and poor-initialized condition.
Furthermore, we show by utilizing the previous converged parameters' locations, the partial persistent training boosts the performance on a range of models. 
Our empirical results on LeNet-5, AlexNet and ResNet-18 show persistent training converges to better solution.
We also show that persistent neurons alleviates the saturation problem.
To our knowledge, this is the fist trajectory-based informed bias method.
Persistent neurons presents a new approach to address some of the main concerns and limitations of landscape conjecture and be easily generalized to more learning tasks.






\bibliographystyle{plain}

\end{document}